\newcommand{\methodname}{\textbf{SDB}}
\newcommand{\longmethodname}{\textbf{Structured Diffusion Bridge}}

\documentclass{article}

\usepackage[table]{xcolor}
\usepackage{microtype}
\usepackage{graphicx}
\usepackage{subcaption}
\usepackage{booktabs} 
\usepackage{pgfplots}
\usepackage{pgfplotstable}
\usepackage{multirow}
\pgfplotsset{compat=1.18} 
\usepgfplotslibrary{groupplots}

\usepackage{hyperref}
\usepackage{array}
\usepackage[most]{tcolorbox}

\usepackage{dsfont}

\usepackage[accepted]{icml2026}

\usepackage{amsmath}
\usepackage{amssymb}
\usepackage{mathtools}
\usepackage{amsthm}
\usepackage{svg}

\usepackage[capitalize,noabbrev]{cleveref}

\theoremstyle{plain}

\theoremstyle{definition}

\theoremstyle{remark}

\usepackage[textsize=tiny]{todonotes}

\usepackage{tikz}
\usetikzlibrary{arrows.meta,positioning,calc,fit}

\icmltitlerunning{Structured Diffusion Bridges: Inductive Bias for Denoising Diffusion Bridges}

\begin{document}

\twocolumn[
  \icmltitle{Structured Diffusion Bridges: Inductive Bias for Denoising Diffusion Bridges}



  \icmlsetsymbol{equal}{*}

  \begin{icmlauthorlist}
    \icmlauthor{Eitan Kosman}{1}
    \icmlauthor{Gabriele Serussi}{1}
    \icmlauthor{Chaim Baskin}{1}
  \end{icmlauthorlist}

  \icmlaffiliation{1}{Ben-Gurion University of the Negev, Israel}

  \icmlkeywords{Machine Learning, ICML}

  \vskip 0.1in
]



\icmlcorrespondingauthor{Eitan Kosman}{kosmane@post.bgu.ac.il}
\printAffiliationsAndNotice{}  

\begin{abstract}
Modality translation is inherently under-constrained, as multiple cross-modal mappings may yield the same marginals. Recent work has shown that diffusion bridges are effective for this task. However, most existing approaches rely on fully paired datasets, thereby imposing a single data-driven constraint. We propose a diffusion-bridge framework that characterizes the space of admissible solutions and restricts it via alignment constraints, treating paired supervision as an optional heuristic rather than a prerequisite. We validate our method on synthetic and real modality translation benchmarks across unpaired, semi-paired, and paired regimes, showing consistent performance across supervision levels. Notably, \textbf{it achieves near fully-paired quality with a substantial relaxation in pairing requirements, and remaining applicable in the unpaired regime}. These results highlight diffusion bridges as a flexible foundation for modality translation beyond fully paired data.
\end{abstract}

\section{Introduction}
\label{sec:introduction}
\begin{figure}[t]
  \centering

  \includegraphics[width=\linewidth]{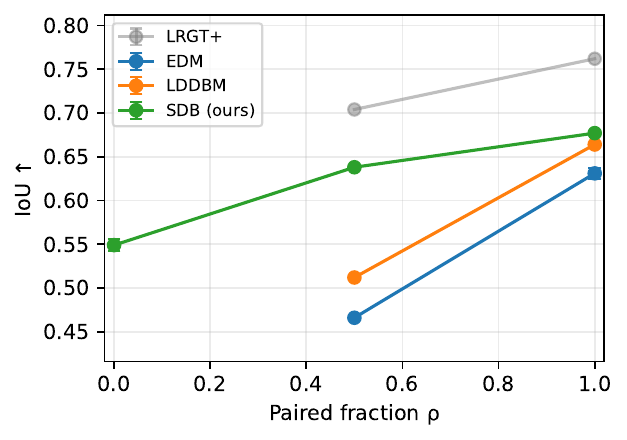}  
  \caption{\textbf{Multi-view $\rightarrow$ 3D (ShapeNet).} Reconstruction fidelity (IoU) versus paired fraction $\rho$ when translating four-view images into $32^3$ voxel grids. \textbf{SDB (ours)} outperforms diffusion-bridge baselines (EDM \cite{karras2022elucidating}, LDDBM \cite{berman2025towards}) whenever paired supervision is available ($\rho \in \{0.5,1.0\}$), while remaining applicable in the fully unpaired setting ($\rho=0$) where paired objectives cannot be trained. LRGT+ \cite{yang2023longrangegroupingtransformermultiview} is an ad-hoc pipeline included as a reference point rather than a principled diffusion-bridge baseline.}
  \label{fig:teaser}
\end{figure}
Mapping information across heterogeneous domains is a central problem in machine learning. Diffusion models have achieved significant success in generative tasks~\citep{sohl2015deep,ho2020denoising,song2020score, berman2024reviving}, and recent advancements have extended these models to distribution-to-distribution translation using diffusion bridges~\citep{de2021diffusion,shi2023diffusion,zhou2024denoising,liu2023isb}. Notably, Latent Denoising Diffusion Bridge Models (LDDBM)~\citep{berman2025towards} have achieved state-of-the-art results by leveraging a shared latent space to address dimensional mismatches. Despite these advances, many bridge approaches predominantly depend on paired correspondences to define and learn effective translations.

The exclusive reliance on paired supervision presents a significant limitation. In conventional denoising diffusion bridge models \cite{zhou2024denoising}, training with paired data implicitly enforces three requirements: semantic correspondence (mapping the source to the target), distributional validity (ensuring that outputs match the target marginal), and geometric consistency (maintaining invertibility). Relying solely on paired loss to learn transport geometry from the ground up may not fully exploit the problem's inherent structure. This strategy can be suboptimal even with abundant data, as the model may minimize reconstruction error while failing to adhere to the underlying manifold. Distinguishing these constraints enables paired supervision to focus on correspondences, while dedicated objectives enforce distributional and geometric properties.

To address this, we introduce \longmethodname ~(\methodname) as a framework that augments LDDBM~\cite{berman2025towards} with composable structural constraints. Instead of relying exclusively on paired data to infer structure, this approach explicitly enforces it: marginal matching anchors the bridge to the target distribution, endpoint cycle consistency promotes invertibility ($y \to x \to \hat{y} \approx y$), and trajectory-level cycle consistency regularizes the stochastic flow. These constraints are independent of paired supervision, enabling the direct incorporation of inductive bias into the transport plan. This formulation advances learning by combining principled geometric enforcement with example-based methods.

Empirical results demonstrate that explicit structural constraints enhance both performance and robustness. In the fully paired setting, \methodname ~surpasses state-of-the-art performance, increasing PSNR on super-resolution (FFHQ$\to$CelebA-HQ) from $25.6$ to $25.9$. Importantly, \methodname ~maintains strong performance as data availability decreases. When supervision is reduced to $50\%$, performance remains competitive with fully supervised baselines. Even in the absence of paired data, structural constraints alone enable meaningful translation ($19.0$ PSNR on super resolution). This pattern of degradation suggests that geometric constraints effectively capture the essential mechanics of translation, implying that dense pairing is not always necessary for semantic alignment.

\textbf{We summarize our contributions as follows}:
\begin{itemize}
    \item We formulate diffusion bridge training as a composite objective, treating paired supervision as one of several constraints.
    
    \item We introduce explicit structural regularization through marginal matching and multi-level cycle consistency, demonstrating that geometric enforcement improves fully paired translation.
    
    \item We demonstrate that \methodname~retains efficacy as paired data decreases, indicating strong performance in low-resource settings.
    
    \item We provide a comprehensive ablation analysis that isolates the contributions of each constraint, highlighting their roles in shaping the diffusion trajectory.
\end{itemize}

\section{Related Work}
\label{sec:related}
\paragraph{Diffusion bridges for modality translation.}
Diffusion bridges have become a strong paradigm for \emph{distribution-to-distribution} translation with diffusion models, including practical instantiations for translation tasks and latent variants such as LDDBM \citep{de2021diffusion,shi2023diffusion,liu2023isb,zhou2024denoising,berman2025towards}.
In most translation-oriented bridge methods, \emph{paired} correspondences are the primary source of identifiability and the main way to enforce semantic alignment, so performance is tightly coupled to dense pairing.
By contrast, classical unpaired translation methods rely on surrogate alignment signals such as cycle consistency or contrastive objectives \citep{zhu2017unpaired,liu2017unsupervised,park2020contrastive}.
We find LADB \citep{wang2025ladb} as a closely related approach. It reduces pairing demands by reusing a pretrained source latent diffusion model to map source samples to latent endpoints and training a target latent diffusion model with a mixture of paired and unpaired latent--target couplings.
Rather than aiming to replace unpaired pipelines or relying on dense pairing, we strengthen bridge-based translation \emph{even at full supervision} via explicit structural constraints, and show they remain informative as pairing becomes scarce, yielding graceful degradation.

\paragraph{Schr\"odinger bridges, optimal transport, and unpaired couplings.}
Schr\"odinger bridges (SB) connect stochastic transport to entropy-regularized optimal transport and provide a principled view of learning couplings between endpoint marginals \citep{de2021diffusion,shi2023diffusion,tong2024simulation}.
This perspective naturally accommodates \emph{unpaired} data, but in modality translation the coupling can remain underdetermined without additional inductive structure.
Recent unpaired SB instantiations illustrate complementary trade-offs: UNSB \citep{kim2024unpaired} casts learning as a sequential adversarial procedure, which can be susceptible to GAN-style failure modes (e.g., instability, mode collapse, and hyperparameter sensitivity), particularly in complex modality translation, while simulation-free approximations depend on a chosen ground cost or feature space that may not match semantic correspondence \citep{tong2024simulation}.
We therefore focus on constraints that improve bridge-based translation under full pairing and remain informative as pairing diminishes.

\section{Background}
\label{sec:background}

\subsection{Diffusion Models}

Diffusion models~\citep{sohl2015deep,ho2020denoising,song2020score} define a forward noising process that gradually perturbs data into a simple prior distribution, typically an isotropic Gaussian. A neural network learns the corresponding reverse-time denoising process via score matching, enabling generation by sampling from the prior and iteratively denoising. While highly expressive, standard diffusion is inherently \emph{prior-to-data}: generation is anchored to a fixed simple distribution, making direct translation between two arbitrary and potentially complex data distributions nontrivial without additional structure.

\subsection{Denoising Diffusion Bridge Models}

Denoising Diffusion Bridge Models (DDBMs)~\citep{zhou2024denoising} extend diffusion to connect two arbitrary endpoint distributions by conditioning on fixed endpoints via Doob's h-transform~\citep{doob1984classical}. Given marginals $p(x_0)$, $p(x_T)$ and a joint distribution $p(x_0, x_T)$, the forward process starts at $x_0$ and is conditioned to terminate at $x_T$:
\begin{equation}
    dx_t = f(x_t, t) \, dt + g(t)^2 h(x_t, t, x_T, T) \, dt + g(t) \, dw_t,
\end{equation}
where $f$, $g$, and $h$ are analytically specified under standard noise schedules. DDBM learns the reverse-time process:
\begin{equation}
\label{eq:reverse_sde}
\begin{aligned}
dx_t
&= \Bigl[ f(x_t, t)
 - g^2(t)\Bigl( \nabla_{x_t}\log q(x_t \mid x_T) \\
&\qquad\qquad\;\; - h(x_t, t, x_T, T) \Bigr) \Bigr]\, dt + g(t)\, d\hat{w}_t .
\end{aligned}
\end{equation}
where $\nabla_{x_t} \log q(x_t | x_T)$ is the score function, approximated by a neural network $s_\theta(x_t, x_T, t)$. Training minimizes a weighted score matching objective: 
\begin{equation}
\label{eq:backward}
\begin{aligned}
\mathcal{L}_{\text{DDBM}}
&= \mathbb{E}_{x_t, x_0, x_T, t}\Bigl[
w(t)\,\bigl\| s_\theta(x_t, x_T, t) \\
&\qquad\qquad\quad
 - \nabla_{x_t}\log q(x_t \mid x_0, x_T) \bigr\|^2
\Bigr].
\end{aligned}
\end{equation}
Under the variance-exploding (VE) scheme, $\nabla_{x_t} \log q(x_t | x_0, x_T) = (x_T - x_t) / (\sigma_T^2 - \sigma_t^2)$ with $x_t \sim \mathcal{N}(x_0, \sigma_t^2 I)$. Reversing the learned bridge yields an approximate sampler for $p(x_0 | x_T)$, enabling translation from $p(x_T)$ to $p(x_0)$. Critically, training requires samples from the joint $p(x_0, x_T)$—i.e., paired correspondences—and assumes both endpoints reside in the same ambient space $\mathbb{R}^d$.

\subsection{Latent Denoising Diffusion Bridge Models}

Latent Denoising Diffusion Bridge Models (LDDBM)~\citep{berman2025towards} address the dimensionality constraint by operating in a shared latent space. Modality-specific encoders $E_X$, $E_Y$ map inputs to latent representations $z_0 = E_X(x)$ and $z_T = E_Y(y)$, while decoders $D_X$, $D_Y$ reconstruct outputs. The diffusion bridge is learned between latent endpoints, enabling translation across modalities of different dimensionality.

LDDBM combines three losses, each requiring paired samples $(x, y)$:
\begin{equation}
    \mathcal{L}_{\text{LDDBM}} = \mathcal{L}_{\text{bridge}} + \mathcal{L}_{\text{pred}} + \mathcal{L}_{\text{InfoNCE}}.
\end{equation}
The bridge loss $\mathcal{L}_{\text{bridge}}$ applies score matching in latent space as in DDBM. The predictive loss $\mathcal{L}_{\text{pred}} = d(D_X \circ B \circ E_Y(y), x)$ enforces end-to-end reconstruction through the full encode-bridge-decode pipeline, where $B$ denotes the bridge and $d$ is a distance metric. The contrastive loss $\mathcal{L}_{\text{infoNCE}}$ aligns latent representations using InfoNCE~\citep{oord2018representation}.

All three components rely on paired correspondences: the bridge loss uses $(z_0, z_T)$ endpoint pairs, the contrastive loss forms positives from paired samples, and the predictive loss matches decoded outputs to paired targets. Paired supervision serves as the sole alignment signal, defining endpoint connections, distributional structure, and geometric preservation. This raises the question: can explicit structural constraints enhance pairing to improve translation, and how does performance change when pairing is limited or missing?

\begin{figure*}[!t]
    \centering
    \includegraphics[width=0.95\linewidth]{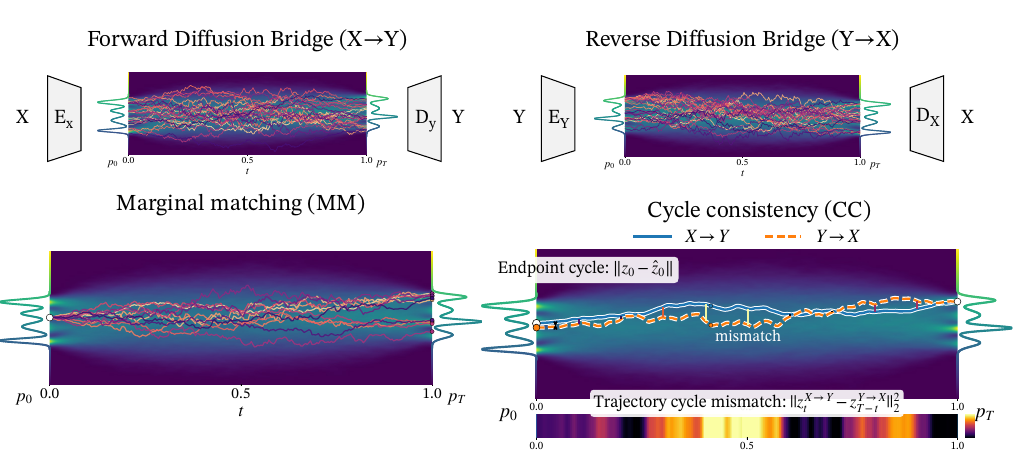}
    \caption{\textbf{Top:} We model unpaired translation via stochastic latent diffusion bridges: a \textit{forward bridge} ($X \to Y$, left) and a \textit{reverse bridge} ($Y \to X$, right). \textbf{Bottom:} Two alignment heuristics. \textit{Marginal Matching (MM)} enforces the terminal constraint $z_T \sim p_Y$. The displayed trajectories illustrate how bridge stochasticity is leveraged to cover the target support, penalizing systematic deviations like mode dropping. \textit{Cycle Consistency (CC)} enforces reversibility ($y \to \hat x \to \hat y \approx y$) by minimizing both the endpoint error $\|z_0-\hat z_0\|$ and the trajectory mismatch $\|z_t^{X\rightarrow Y}-z_{T-t}^{Y\rightarrow X}\|$, visualized in the time-indexed heatmap.}
    \label{fig:method_overview}
\end{figure*}

\section{Heuristic Composition with Latent Denoising Diffusion Bridges}
\label{sec:heuristic_composition}

We model modality translation (MT) as the problem of learning a stochastic diffusion bridge between two marginal distributions \( p_{\mathcal{X}} \) and \( p_{\mathcal{Y}} \) in a shared latent space (Figure~\ref{fig:method_overview}). Following the Latent Denoising Diffusion Bridge Model (LDDBM)~\cite{berman2025towards}, we first encode samples from each modality into a common latent representation, and perform translation by sampling from a denoising diffusion process conditioned on the source modality and constrained to end at the target marginal.

When only the marginals \( p_{\mathcal{X}} \) and \( p_{\mathcal{Y}} \) are provided, modality translation is inherently under-constrained: infinitely many stochastic processes can interpolate between the same endpoint distributions. Equivalently, the set of admissible joint distributions is
\begin{equation}
\label{eq:P}
\mathcal{P}=\{\,p(x,y)\mid p(x)=p_{\mathcal{X}},\;p(y)=p_{\mathcal{Y}}\,\},
\end{equation}
and marginal information alone does not identify a unique coupling between modalities. We therefore view MT as requiring the imposition of \emph{heuristics} that restrict the geometry, stochasticity, and reversibility of the learned diffusion bridge.

Under this perspective, the joint distribution \( p(x,y) \) represented by a paired dataset corresponds to one possible data-driven heuristic for identifying a valid bridge via Doob’s \(h\)-transform~\cite{doob1984classical}, as employed in denoising diffusion bridge models~\cite{berman2025towards, zhou2024denoising}, but is not a prerequisite for learning meaningful cross-modal mappings. In contrast, diffusion bridges provide a foundation for the composition of alternative heuristics, as their diffusion-based formulation exposes intermediate states, time-dependent score fields, and stochastic trajectories that can be directly regularized (Figure~\ref{fig:method_overview}). For generality, we adopt Latent Denoising Diffusion Bridge Models (LDDBMs)~\cite{berman2025towards}, which are not constrained by endpoint dimensionality and enable evaluation across diverse modalities and tasks.

\paragraph{Paired supervision as a heuristic.}
Our formulation also encompasses the original data-driven heuristic of denoising diffusion bridge models (DDBMs) \cite{zhou2024denoising, berman2025towards}, which constrains the diffusion bridge using paired samples. When the dataset is \emph{semi-paired}, meaning that the joint distribution \( p(x, y) \) is known only for a subset of the data, we can apply paired supervision selectively to that subset. The resulting objective naturally combines paired likelihood-based constraints with unpaired heuristics, which are the focus of the discussion that follows.

\paragraph{Pretrained representations as structural priors.}
In addition to the heuristics listed below, pretrained autoencoders and foundation models offer a natural source of inductive bias. By generating a latent parameterization that reflects modality-specific structure, diffusion bridges operate on representations that already align with intrinsic geometry. This condensed form reduces the need for the bridge to identify low-level regularities independently. As demonstrated in our experiments, constraints such as bridge reversibility become more meaningful when applied in the latent space.

\subsection{Marginal Matching}
\label{sec:marginal_matching}
Endpoint Marginal Matching (MM) constrains the diffusion bridge to terminate at the target marginal distribution \( p_{\mathcal{X}} \). During training, we draw \text{\( x \sim p_{\mathcal{X}} \)} and \text{\( y \sim p_{\mathcal{Y}} \)}. Target samples are encoded into \text{\( z_0 = E_{\mathcal{X}}(x) \)} and corrupted using a known forward noising process. Then, we train the diffusion bridge score function \text{\( s_\theta(z_t,t \mid y) \)} via denoising score matching (DSM) by minimizing
\[
\mathcal{L}_{\mathrm{DSM}}
=
\mathbb{E}_{z_0,\,t,\,z_t}
\left[
\big\|
s_\theta(z_t,t \mid y)
-
\nabla_{z_t}\log q(z_t \mid z_0)
\big\|_2^2
\right].
\]

To reduce coupling ambiguity, we employ a winner-takes-all (WTA) assignment heuristic. For each target sample \( z_0 \), multiple conditioning candidates \( \{y^{(k)}\}_{k=1}^K \sim p_{\mathcal{Y}} \) are drawn. We evaluate \( \mathcal{L}_{DSM} \) and propagate gradients only through the candidate
\[
k^\star = \arg\min_{k} \ \mathcal{L}_{DSM}(z_0, y^{(k)}),
\]
which selects the condition \(y^{k^*}\) that best explains the target sample with respect to the learned bridge.

Yet, WTA may lead to \emph{condition dominance}, where a few \emph{low-information} conditioning samples \text{\( \mathcal{S}\subset\{y^{(i)}\}_{i=1}^{N} \)} are selected for a relatively large number of target samples. We alleviate this by imposing a capacity constraint \text{\( C_{y}=2 \)} that limits the number of selections of each conditioning sample \( y^{(i)} \) within an epoch.

The WTA assignment is an optimization heuristic for reducing marginal-coupling ambiguity, not an identifiability guarantee. Its role is to select, among candidate conditioning endpoints, the one that is locally most compatible with the current bridge score. The capacity constraint prevents a small number of low-information conditioning samples from dominating this assignment. In Appendix \ref{app:capacity_sensitivity}, we report sensitivity to the capacity value and to the number of WTA candidates, showing that moderate capacity improves coupling quality while additional candidates provide a compute–accuracy trade-off with diminishing returns.

\subsection{Cycle Consistency as Bridge Reversibility}
\label{sec:cycle_reversibility}

We propose a Cycle-Consistency (CC) heuristic based on a \emph{bridge reversibility constraint}, which limits admissible diffusion bridges to those that are approximately invertible at the level of latent stochastic flows. In unpaired settings, marginal matching alone can lead to degenerate couplings that mix source and target samples in arbitrary ways. Reversibility enforcement helps reduce this ambiguity by penalizing irreversible information loss. We formalize reversibility at endpoints, but also along intermediate trajectories, thereby imposing a strong constraint on the internal structure of the stochastic process. In contrast to classical cycle consistency methods based on deterministic reconstruction \cite{zhu2017unpaired}, our approach enforces consistency between forward and reverse diffusion bridges \cite{berman2025towards, zhou2024denoising}. This adaptation makes the method naturally compatible with LDDBM and more robust to the inherent stochasticity of diffusion models.

\paragraph{\textbf{Forward and reverse diffusion bridges.}}
Let $s_{\mathcal{X}\rightarrow\mathcal{Y}}(z,t)$ denote the score function of the diffusion bridge translating from modality $\mathcal{X}$ to $\mathcal{Y}$, and $s_{\mathcal{Y}\rightarrow\mathcal{X}}(z,t)$ the corresponding reverse bridge. Given a latent source sample $z_0 \sim p_{\mathcal{X}}$, a forward bridge generates
\begin{equation}
z_T = \Phi_{\mathcal{X}\rightarrow\mathcal{Y}}(z_0),
\end{equation}
where $\Phi$ denotes the stochastic flow induced by the diffusion process. Applying the reverse bridge yields
\begin{equation}
\hat z_0 = \Phi_{\mathcal{Y}\rightarrow\mathcal{X}}(z_T).
\end{equation}
Bridge reversibility requires $\hat z_0$ to be statistically consistent with the original latent $z_0$.

\paragraph{\textbf{Latent endpoint reversibility.}}
The simplest instantiation of cycle consistency enforces agreement between the original and reconstructed latent representations:
\begin{equation}
\mathcal{L}_{\text{cycle}}^{\text{end}} =
\mathbb{E}_{z_0 \sim p_{\mathcal{X}}}
\big[
\| \hat z_0 - z_0 \|_2^2
\big].
\end{equation}
This constraint encourages approximate invertibility at the endpoints of the diffusion bridge, while allowing stochasticity along the trajectory.

\paragraph{\textbf{Trajectory-level reversibility.}}
Endpoint consistency alone is insufficient to constrain the internal structure of a diffusion bridge. We therefore extend cycle consistency to intermediate diffusion states along the trajectory. Let $\{z_t^{X \rightarrow Y}\}_{t=0}^T$ denote the latent states of the forward bridge, and $\{z_{T-t}^{Y \rightarrow X}\}_{t=0}^T$ the corresponding states of the reverse bridge. We enforce Trajectory-level reversibility by requiring that these forward and reverse latent paths remain approximately consistent at matching diffusion times, via
\begin{equation}
\mathcal{L}_{\text{cycle}}^{\text{traj}} =
\mathbb{E}_{t,z_t}
\Big[
w(t)\,
\| z_t^{X\rightarrow Y} - z_{T-t}^{Y\rightarrow X} \|_2^2
\Big],
\quad
w(t)=\frac{1}{\sigma_t^2+\varepsilon}.
\end{equation}
This enforces reversibility across diffusion time while accounting for the scale variation.

\subsection{Overall Objective}
\label{sec:overall_objective}

The final training objective composes the heuristic constraints. We assume access to datasets \text{$\mathcal{D_{\mathcal{X}}}=\{x_i\}_{i=1}^{N_{\mathcal{X}}}$} and \text{$\mathcal{D_{\mathcal{Y}}}=\{y_i\}_{i=1}^{N_{\mathcal{Y}}}$}, which define the empirical marginal distributions $p_{\mathcal{X}}$ and $p_{\mathcal{Y}}$ over modalities $\mathcal{X}$ and $\mathcal{Y}$, respectively. In the semi-paired regime, we additionally observe a paired subset \text{$\mathcal{D}_{\mathrm{pair}}\subseteq \{(x_i,y_j): x_i\in\mathcal{D}_{\mathcal{X}},\,y_j\in\mathcal{D}_{\mathcal{Y}}\}$},
i.e., a set of samples $(x,y)$ with $x\sim p_{\mathcal{X}}$ and $y\sim p_{\mathcal{Y}}$ for which a correspondence is available. We optimize the unified objective
\begin{equation}
\begin{aligned}
\mathcal{L}_{\mathrm{total}}
&= \mathcal{L}_{\mathrm{DSM}}
+ \lambda_{\mathrm{end}}\mathcal{L}_{\mathrm{cycle}}^{\mathrm{end}}
+ \lambda_{\mathrm{traj}}\mathcal{L}_{\mathrm{cycle}}^{\mathrm{traj}} \\
&\quad + \lambda_{\mathrm{pair}}\,\mathds{1}_{(x,y)\in\mathcal{D}_{\mathrm{pair}}}\mathcal{L}_{\mathrm{pair}} .
\end{aligned}
\end{equation}
where $\mathcal{L}_{\mathrm{DSM}}$ enforces termination at the target marginal (Sec.~\ref{sec:marginal_matching}) and $\mathcal{L}_{\mathrm{cycle}}^{\mathrm{end}},\mathcal{L}_{\mathrm{cycle}}^{\mathrm{traj}}$ impose bridge reversibility at endpoints and along trajectories (Sec.~\ref{sec:cycle_reversibility}). The final term recovers the original data-driven heuristic of DDBMs \cite{zhou2024denoising, berman2025towards} by applying paired supervision $\mathcal{L}_{pair}$ only on $\mathcal{D}_{\mathrm{pair}}$. We use $\lambda=1$ for all objectives, as we have not observed significant improvement with other values. This unified formulation composes multiple heuristics to select a diffusion-bridge-induced joint $p_\theta(x,y)\in\mathcal{P}$ (Eq.~\ref{eq:P}) that is consistent with the marginals while being biased toward reversible, condition-preserving transport. Appendix~\ref{sec:heuristic_limitations} summarizes the complementary role of each heuristic and the failure mode it addresses.

\begin{figure*}[t]
\begin{subfigure}[t]{0.13\textwidth}
\centering
\begin{tikzpicture}

\pgfplotstableread[col sep=space]{
rho paired_acc ours_acc
0.0 0.183000 0.868333
0.1 0.355000 0.883667
0.5 0.641000 0.954667
1.0 0.887333 0.965000
}\acc

\begin{axis}[
    height=2.6cm,
    scale only axis,
    grid=both,
    xlabel={$\rho$},
    ylabel={\shortstack{Content acc. \\$\rightarrow$}},
    ylabel style={at={(axis description cs:-0.15,0.5)},anchor=south},
    xmin=0, xmax=1,
    ymin=0.1, ymax=1.0,
    title={Coupling quality},
    tick label style={font=\scriptsize},
]
\addplot+[
  mark=o,
  x filter/.code={
    \pgfmathparse{\pgfmathresult==0 ? nan : \pgfmathresult}%
  },
] table[x=rho, y=paired_acc]{\acc};
\addplot+[mark=square] table[x=rho, y=ours_acc]{\acc};
\end{axis}
\end{tikzpicture}
\end{subfigure}\hspace{0.11\textwidth}
\begin{subfigure}[t]{0.13\textwidth}
\centering
\begin{tikzpicture}
\pgfplotstableread[col sep=space]{
rho paired_cycle ours_cycle
0.0 0.977830 0.679605
0.1 0.957288 0.668844
0.5 0.844565 0.615000
1.0 0.685644 0.603987
}\cyc

\begin{axis}[
    height=2.6cm,
    scale only axis,
    grid=both,
    xlabel={$\rho$},
    ylabel={\shortstack{Cycle MSE \\$\leftarrow$}},
    ylabel style={at={(axis description cs:-0.15,0.5)},anchor=south},
    xmin=0, xmax=1,
    title={Reversibility},
    tick label style={font=\scriptsize},
]
\addplot+[
  mark=o,
  x filter/.code={
    \pgfmathparse{\pgfmathresult==0 ? nan : \pgfmathresult}%
  },
] table[x=rho, y=paired_cycle]{\cyc};
\addplot+[mark=square] table[x=rho, y=ours_cycle]{\cyc};
\end{axis}
\end{tikzpicture}
\end{subfigure}\hspace{0.12\textwidth}
\begin{subfigure}[t]{0.13\textwidth}
\centering
\begin{tikzpicture}
\pgfplotstableread[col sep=space]{
rho paired_swd ours_swd
0.0 0.021336 0.019676
0.1 0.022075 0.021102
0.5 0.019514 0.019300
1.0 0.023158 0.019372
}\swd

\begin{axis}[
    height=2.6cm,
    scale only axis,
    grid=both,
    xlabel={$\rho$},
    ylabel={\shortstack{SWD \\$\leftarrow$}},
    ylabel style={at={(axis description cs:-0.05,0.5)},anchor=south},
    xmin=0, xmax=1,
    title={Marginal alignment (SWD)},
    tick label style={font=\scriptsize},
]
\addplot+[
mark=o,
x filter/.code={
\pgfmathparse{\pgfmathresult==0 ? nan : \pgfmathresult}%
},
] table[x=rho, y=paired_swd]{\swd};
\addplot+[mark=square] table[x=rho, y=ours_swd]{\swd};
\end{axis}
\end{tikzpicture}
\end{subfigure}\hspace{0.1\textwidth}
\begin{subfigure}[t]{0.13\textwidth}
\centering
\begin{tikzpicture}
\pgfplotstableread[col sep=space]{
rho paired_mmd ours_mmd
0.0 -0.000057 -0.000111
0.1 -0.000037 -0.000093
0.5 -0.000133 -0.000151
1.0 -0.000019 -0.000147
}\mmd

\begin{axis}[
    height=2.6cm,
    scale only axis,
    grid=both,
    xlabel={$\rho$},
    ylabel={\shortstack{MMD$^2$ \\$\leftarrow$}},
    ylabel style={at={(axis description cs:-0.1,0.5)},anchor=south},
    xmin=0, xmax=1,
    title={Marginal alignment (MMD$^2$)},
    tick label style={font=\scriptsize},
]
\addplot+[
mark=o,
x filter/.code={
\pgfmathparse{\pgfmathresult==0 ? nan : \pgfmathresult}%
},
] table[x=rho, y=paired_mmd]{\mmd};
\addplot+[mark=square] table[x=rho, y=ours_mmd]{\mmd};
\end{axis}
\end{tikzpicture}
\end{subfigure}
\begin{center}
\vspace{-1.5em}
\begin{tikzpicture}
\begin{axis}[
  hide axis,
  xmin=0, xmax=1,
  ymin=0, ymax=1,
  legend columns=2,
  legend style={
    draw=black,          
    line width=0.4pt,    
    fill=white,          
    fill opacity=0.9,    
    text opacity=1,      
    rounded corners=5pt, 
    inner sep=2pt,       
    column sep=1.5em,
    at={(0.5,0.5)},
    anchor=center,
  },
]
\addlegendimage{blue,mark=o}
\addlegendentry{Paired-only}

\addlegendimage{red,mark=square}
\addlegendentry{Semi-paired (ours)}
\end{axis}
\end{tikzpicture}
\vspace{-0.5em}
\end{center}
\captionsetup{justification=raggedright,singlelinecheck=false}
\caption{\textbf{Synthetic benchmark results across supervision levels (\(\rho\))}. Semi-paired training consistently improves coupling quality and cycle consistency relative to paired-only baselines, while maintaining competitive target marginal alignment.}
\label{fig:synthetic_rho_grid}
\end{figure*}
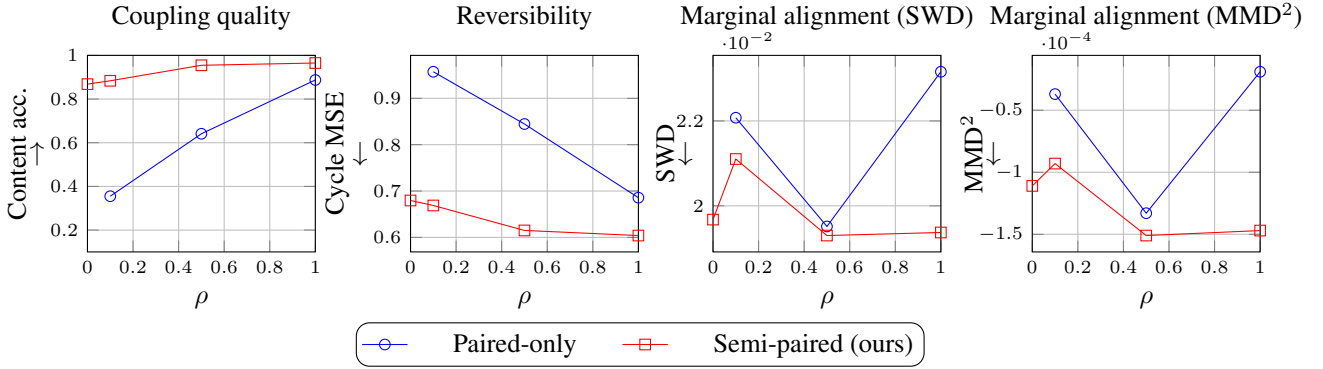

\begin{table*}[h]
\centering
\caption{\textbf{Effect of structured constraints across supervision regimes.}
Endpoint and trajectory reversibility consistently improve coupling quality and cycle consistency across $\rho=0,0.5,1$ settings, while maintaining competitive marginal alignment.}
\label{tab:heuristic_ablation}
\begin{tabular}{llcccc}
\toprule
$\rho$ & Method
& SWD $\downarrow$
& MMD$^2$ $\downarrow$
& Content Acc.\ $\uparrow$
& Cycle MSE $\downarrow$ \\
\midrule
\multirow{3}{*}{0}
& Marginal matching only
& 0.02021
& $-1.03\times10^{-4}$
& 0.162
& 0.972 \\
& + Endpoint cycle
& \textbf{0.01891}
& $-1.69\times10^{-4}$
& 0.662
& 0.831 \\
& + Trajectory cycle
& 0.019676
& $-1.11\times10^{-4}$
& 0.868333
& 0.679605 \\
\midrule
\multirow{5}{*}{0.5}
& Marginal matching only
& 0.02031
& $-9.5\times10^{-5}$
& 0.167
& 0.991 \\
& + Endpoint cycle
& 0.01918
& $\boldsymbol{-1.72\times10^{-4}}$
& 0.645
& 0.824 \\
& + Trajectory cycle
& 0.02206
& $-5.5\times10^{-5}$
& 0.789
& 0.737 \\
& Paired-only
& 0.01951
& $-1.33\times10^{-4}$
& 0.641
& 0.845 \\
& Semi-paired (ours)
& 0.01930
& $-1.51\times10^{-4}$
& 0.955
& 0.615 \\
\midrule
\multirow{5}{*}{1.0}
& Marginal matching only
& 0.01915
& $-1.47\times10^{-4}$
& 0.178
& 0.976 \\
& + Endpoint cycle
& 0.02289
& $-2.0\times10^{-6}$
& 0.671
& 0.830 \\
& + Trajectory cycle
& 0.01949
& $-1.37\times10^{-4}$
& 0.768
& 0.753 \\
& Paired-only
& 0.02316
& $-1.9\times10^{-5}$
& 0.887
& 0.686 \\
& Semi-paired (ours)
& 0.01937
& $-1.47\times10^{-4}$
& \textbf{0.965}
& \textbf{0.604} \\
\bottomrule
\end{tabular}
\end{table*}
\section{Experiments}
\label{sec:experiments}
We validated \methodname{} across a range of translation tasks, including a controlled synthetic dataset in Section~\ref{exp:synthetic} and real-world scenarios in Section~\ref{exp:sr} and Section~\ref{exp:mv2vox}. The objectives of this empirical study were fourfold:
\begin{enumerate}
    \item 
    How do the proposed heuristics affect performance within our framework?
    \item 
    As we first and foremost target improving general modality translation, how well does \methodname{} perform under common evaluation protocols?
    \item 
    We further demonstrate the superiority of \methodname{} in unpaired and semi-paired regimes. Can structured bridges learn meaningful coupling via constraining properties of stochastic flows?
    \item 
    Can we attribute high-quality coupling to strong pretrained encoders that induce modality-native latent parameterization
\end{enumerate}
 We adopt the same tasks and architectural components as in prior works \cite{berman2025towards, wu20153d, moser2024waving, rombach2021highresolution, karras2022elucidating} to enable direct comparison while systematically relaxing the availability of paired data. For all tasks, we implement the diffusion bridge in both directions (\(\mathcal{X}\!\to\!\mathcal{Y}\) and \(\mathcal{Y}\!\to\!\mathcal{X}\)) since cycle-consistency terms require translating outputs back to the source modality. Unless otherwise stated, changing the paired fraction $\rho$ changes only which
samples have correspondence labels; the total number of endpoint samples used
to estimate the two marginals is held fixed. Thus, decreasing $\rho$ evaluates
weaker correspondence supervision rather than simply reducing the amount of
marginal data.

\subsection{Synthetic Benchmark of Direct Endpoint Bridges}
\label{exp:synthetic}

We begin with a controlled synthetic setting that allows quantitative evaluation of coupling quality at various levels of supervision. Unlike real-data experiments that rely on modality-specific encoders and decoders, we consider direct endpoint bridges and operate in a shared latent space, using a DDBM-style formulation \cite{zhou2024denoising}.

We generate samples \(z_0 \sim p_{\mathcal{X}}\) and \(z_T \sim p_{\mathcal{Y}}\) from a structured content--style model in which endpoints share a discrete content variable and nuisance style factors induce ambiguity in the marginal coupling. We implement the denoiser as a stack of multi-head attention (MHA) blocks, conditioned on \(y\) and diffusion time.
We refer the reader to Appendix \ref{app:exp:synthetic_data} for a description of the data generation process, the denoiser architecture, the evaluation metrics, and additional results for various data configurations.

\begin{table*}[t]
\centering
\setlength{\tabcolsep}{5.0pt}
\renewcommand{\arraystretch}{1.15}
\caption{
\textbf{Zero-shot SR on CelebA-HQ when training on FFHQ, sweeping paired fraction \(\rho\).}
We report PSNR, SSIM and LPIPS.
DiWa \cite{moser2024waving} and LDDBM \cite{berman2025towards} are not applicable at \(\rho=0\) since their objectives assume paired LR--HR supervision.
For \(\rho=1.0\), DiWa and LDDBM entries are filled by results from the original works and re-evaluated for \(\rho=0.5\).}
\begin{tabular}{l c ccc ccc}
\toprule
& \multicolumn{1}{c}{$\rho=0$}
& \multicolumn{3}{c}{$\rho=0.5$}
& \multicolumn{3}{c}{$\rho=1.0$} \\
\cmidrule(lr){2-2}\cmidrule(lr){3-5}\cmidrule(lr){6-8}
& \methodname
& DiWa & LDDBM & \methodname
& DiWa & LDDBM & \methodname \\
\midrule
PSNR$\uparrow$
& \textbf{19.0$\pm$0.6}
& 22.6$\pm$0.2 & 24.9$\pm$0.3 & \textbf{25.2$\pm$0.3}
& 23.3 & 25.6$\pm$0.4 & \textbf{25.9$\pm$0.3} \\
SSIM$\uparrow$
& \textbf{0.54$\pm$0.05}
& 0.62$\pm$0.02 & 0.67$\pm$0.02 & \textbf{0.68$\pm$0.02}
& 0.65 & 0.68$\pm$0.03 & \textbf{0.69$\pm$0.02} \\
LPIPS$\downarrow$
& \textbf{0.37$\pm$0.04}
& 0.40$\pm$0.01 & 0.33$\pm$0.01 & \textbf{0.32$\pm$0.01}
& 0.39 & 0.32$\pm$0.01 & \textbf{0.31$\pm$0.01} \\
\bottomrule
\end{tabular}
\label{tab:sr_main}
\end{table*}

\paragraph{\textbf{Effect of structured constraints on coupling.}}
Table~\ref{tab:heuristic_ablation} presents the effects of incrementally adding structural constraints across different supervision levels. Figure~\ref{fig:synthetic_rho_grid} depicts how these effects evolve as a function of the supervision parameter \(\rho\). In the fully unpaired scenario (\(\rho=0\)), marginal matching alone provides reasonable alignment of endpoint distributions. This is indicated by low SWD and MMD$^2$ values, which measure agreement between the generated and target endpoint marginals, and therefore diagnose distributional validity rather than source-conditioned correspondence. However, this approach does not learn condition-informative couplings, resulting in poor content accuracy and elevated endpoint cycle error. Introducing endpoint-level cycle consistency substantially reduces cycle MSE and significantly improves content accuracy. This suggests that endpoint reversibility partially addresses coupling ambiguity. Enforcing cycle consistency throughout the entire diffusion trajectory further improves condition preservation in the learned coupling. This underscores the need to constrain intermediate stochastic flows rather than rely solely on endpoint agreement.

The combination of all constraints produces the strongest performance in the unpaired regime, achieving an optimal balance between coupling quality and reversibility while maintaining competitive marginal alignment. This pattern is consistently observed across all \(\rho\) values in Figure~\ref{fig:synthetic_rho_grid}. Compared to paired-only baselines, semi-paired training shows monotonic improvements in coupling quality and reversibility, while marginal alignment metrics remain stable. Overall, these trends indicate that the observed improvements result from enhanced structural alignment between endpoints rather than overfitting to the target distribution.

This trend continues in the semi-paired setting (\(\rho=0.5\)) and extends to the fully paired regime (\(\rho=1.0\)). Semi-paired training with structured constraints consistently outperforms paired-only baselines at all supervision levels, including \(\rho=1.0\). This demonstrates the complementary nature of partial pairing and heuristic regularization. Additional results for alternative synthetic data generation configurations are provided in Appendix~\ref{app:exp:synthetic_data_results}. Collectively, these findings support the conclusion that modality translation via diffusion bridges is fundamentally under-constrained by marginals alone. The process benefits from the explicit integration of structural heuristics that influence the geometry and reversibility of the learned stochastic coupling.


\subsection{Image Super-Resolution}
\label{exp:sr}

Single-image super-resolution (SR) is formulated as a modality translation problem from low-resolution inputs \(y \in \mathcal{Y}\) to high-resolution targets \(x \in \mathcal{X}\). Within the structured diffusion bridge framework, SR involves learning a conditional stochastic coupling that transports mass from the low-resolution endpoint distribution to the high-resolution endpoint distribution, while preserving the conditioning signal provided by \(y\). The same supervision protocol as described in Section~\ref{exp:synthetic} is employed, varying the paired fraction \(\rho \in [0,1]\) while maintaining a fixed total number of training samples. Appendix~\ref{app:exp:sr} contains detailed descriptions of data construction, model architectures, and training hyperparameters.

\paragraph{\textbf{Setup.}}

We map both endpoints to a shared latent space using modality-specific autoencoders. Concretely, we encode LR observations as \(z_T = E_{\mathrm{LR}}(x)\) and HR targets as \(z_0 = E_{\mathrm{HR}}(y)\), and learn a diffusion bridge between these latent endpoints with a denoiser \(s_\theta(z_t, t \mid z_T)\) conditioned on the LR endpoint. We also attempted to bypass autoencoders and learn the bridge directly in image space using a DDBM-style formulation \cite{zhou2024denoising}, but this was unsuccessful for SR. In contrast, the approach worked reliably only with sufficiently strong autoencoders, suggesting that the learned latent representations reveal a more structured geometry in which the LR--HR correspondence is easier to capture. We refer the reader to Appendix~\ref{app:exp:sr} for further details on the evaluation protocol, autoencoders, denoiser, and sampling/training.

\paragraph{\textbf{Baselines.}}

We compare against DiWa~\cite{moser2024waving} and LDDBM~\cite{berman2025towards} because they represent two different forms of paired-supervision baselines for SR. DiWa is a task-specific diffusion-wavelet SR method designed for low-to-high resolution reconstruction, while LDDBM is the closest general latent diffusion-bridge baseline. Both require paired LR–HR supervision and therefore are not applicable at $\rho=0$. This distinction is important: DiWa measures performance against a strong SR-specific inductive bias, whereas LDDBM measures the benefit of adding SDB’s structural bridge constraints to a general bridge-based translation framework.

\begin{table*}[t]
\centering
\setlength{\tabcolsep}{4.0pt}
\renewcommand{\arraystretch}{1.12}
\caption{
\textbf{Multiview-to-3D generation on ShapeNet.} We sweep paired fraction \(\rho\in[0,1]\) and report 1-NNA to assess distributional similarity between generated and real shapes, and IoU to assess reconstruction fidelity.
For \(\rho=1.0\), the LRGT+, \textsc{EDM} and LDDBM entries are filled by results reported in the original works \cite{yang2023longrangegroupingtransformermultiview, karras2022elucidating, berman2025towards}. Results for \(\rho=0.5\) were obtained by running the code 5 times each, and absent for \(\rho=0\) as they require paired data during training.
}
\resizebox{\textwidth}{!}{%
\begin{tabular}{l c cccc cccc}
\toprule
& \multicolumn{1}{c}{$\rho=0$}
& \multicolumn{4}{c}{$\rho=0.5$}
& \multicolumn{4}{c}{$\rho=1.0$} \\
\cmidrule(lr){2-2}\cmidrule(lr){3-6}\cmidrule(lr){7-10}
& \methodname
& \cellcolor{gray!15}{LRGT+} & EDM & LDDBM & \methodname
& \cellcolor{gray!15}{LRGT+} & EDM & LDDBM & \methodname \\
\midrule
1-NNA$\downarrow$
& \textbf{0.657$\pm$0.008}
& \cellcolor{gray!15}{\text{--}} & 0.697$\pm$0.013 & 0.660$\pm$0.008 & \textbf{0.559$\pm$0.007}
& \cellcolor{gray!15}{\text{--}} & 0.532$\pm$0.013 & 0.508$\pm$0.005 & \textbf{0.481$\pm$0.003} \\
IoU$\uparrow$
& \textbf{0.549$\pm$0.007}
& \cellcolor{gray!15}{\text{$0.704\pm0.1$}} & 0.466$\pm$0.006 & 0.512$\pm$0.004 & \textbf{0.638$\pm$0.004}
& \cellcolor{gray!15}{0.762} & 0.631$\pm$0.006 & 0.664$\pm$0.002 & \textbf{0.677$\pm$0.002} \\
\bottomrule
\end{tabular}%
}
\label{tab:mv2vox_main}
\end{table*}
\paragraph{\textbf{Results.}}
Table~\ref{tab:sr_main} shows zero-shot super-resolution results on CelebA-HQ, using models trained on FFHQ (\(16{\times}16 \rightarrow 128{\times}128\)) at different supervision levels \(\rho\). \methodname{} consistently delivers the best overall results, improving both pixel-level accuracy and perceptual quality whenever the baselines can be used ($\rho>0$). In the semi-paired and fully paired settings (\(\rho=0.5,1.0\)), \methodname{} achieves the highest PSNR and SSIM and the lowest LPIPS, which means it reconstructs images more accurately, preserves structure, and looks more realistic under domain shift. \methodname{} also works well in the fully unpaired case (\(\rho=0\)), where DiWa \cite{moser2024waving} and LDDBM \cite{berman2025towards} cannot be used. It achieves competitive perceptual quality, though with more variation between runs. This variability likely comes from the greater uncertainty in unpaired super-resolution, since many high-resolution images can match the same low-resolution input. Still, the strong average performance shows that the structured diffusion bridge can find a meaningful mapping from low- to high-resolution images even without paired data. To verify that this trend is not specific to the coarse $\rho \in \{0,0.5,1.0\}$ sweep in Table~\ref{tab:sr_main}, Appendix~\ref{app:sr_finer_sweep} reports additional intermediate paired fractions while keeping the total number of endpoint samples fixed.

Qualitative results are included in Appendix~\ref{app:sr_results}. We additionally report a component-wise ablation on this benchmark in Appendix~\ref{app:sr_ablation}, showing that the full combination of marginal matching, endpoint cycle consistency, and trajectory-level cycle consistency gives the best overall trade-off across PSNR, SSIM, and LPIPS.

\paragraph{\textbf{Connection to the synthetic benchmark.}}
The SR setting serves as a real-data analogue to the synthetic content--style benchmark described in Section~\ref{exp:synthetic}. In that context, translation is under-constrained because multiple couplings can match endpoint marginals without preserving the shared \emph{discrete} content label. Similarly, SR exhibits this ambiguity, but with \emph{continuous} content distributed over the image manifold. Therefore, effective SR requires a coupling that aligns continuous content across LR and HR endpoints while treating high-frequency details as nuisance variables. This also clarifies a limitation of the approach. \methodname{} relies on latent representations in which cross-modal content is sufficiently exposed for reversibility and marginal constraints to act meaningfully. In raw pixel space, LR content and HR texture are strongly entangled, and we found direct image-space bridges unreliable for SR. The autoencoder latent space therefore functions as a structural prior: it does not by itself determine the coupling, but it provides a representation in which \methodname{}’s bridge constraints can regularize the transport more effectively. We next evaluate whether the same structured-bridge formulation remains effective on a more challenging cross-modal benchmark.


\subsection{Multi-view Images to 3D Voxel Grids}
\label{exp:mv2vox}

Moving beyond super-resolution, we evaluate \methodname{} on the challenging cross-modal task of generating 3D shapes from 2D multi-view images. In this task, the model receives multiple images of the same object from multiple viewpoints and the corresponding 3D shape during training. The goal at inference time is to generate a 3D sample conditioned on multi-view inputs that follows the distribution of real 3D shapes. We follow the same supervision protocol as in Section~\ref{exp:synthetic}, varying the paired fraction \(\rho \in [0,1]\) while keeping the total number of training samples fixed. Detailed descriptions of the dataset, voxelization, architectures, metrics, and baselines are provided in Appendix~\ref{app:exp:mv2vox}.

\paragraph{\textbf{Setup.}}
We follow the ShapeNet protocol established in \cite{shi20213d}. Each object is rendered into four 2D views captured from uniformly spaced azimuth angles around the object, forming the input distribution \(p_{\mathcal{Y}}(y)\), while 3D occupancy grids define the output distribution \(p_{\mathcal{X}}(x)\). Similar to the methodology described in Section \ref{exp:sr}, we map both endpoints to a shared latent space using modality-specific encoders and decoders. Concretely, we encode the multi-view observation as \(z_T = E_{\mathrm{MV}}(y)\), and encode voxel grids as \(z_0 = E_{\mathrm{Vox}}(x)\). We then learn a diffusion bridge between these latent endpoints with a denoiser \(s_\theta(z_t,t\mid z_T)\) conditioned on the multi-view endpoint. At inference time, given only the multi-view input \(y\), we sample the conditional bridge to obtain \(\hat{z}_0\) and decode a voxel prediction \(\hat{x}=D_{\mathrm{Vox}}(\hat{z}_0)\). We refer the reader to Appendix~\ref{app:exp:mv2vox} for further specifications of the evaluation protocol, architectures, and sampling/training details.

\paragraph{\textbf{Results.}}

Table~\ref{tab:mv2vox_main} reports the results on ShapeNet~\cite{wu20153d} across various supervision levels. \methodname{} consistently achieves strong performance in paired or semi-paired regimes, outperforming diffusion-based baselines~\cite{karras2022elucidating, berman2025towards}. While the task-specific LRGT+~\cite{yang2023longrangegroupingtransformermultiview} attains the best results for \(\rho =1\), \methodname{} remains competitive despite being more general.

As we limit supervision, performance degrades for all methods; however, \methodname{} exhibits substantially smaller degradation, indicating improved robustness to limited paired data. We observe a similar trend for LRGT+, which we attribute to its strong inductive bias. In contrast, EDM~\cite{karras2022elucidating} and LDDBM~\cite{berman2025towards} show a pronounced drop in performance as \(\rho\) decreases, suggesting that purely data-driven diffusion models struggle to maintain consistent multiview-to-shape couplings in the absence of explicit regularization or structural constraints. Notably, while baselines are not applicable when \(\rho=0\), ~\methodname{} remains applicable. A component-wise ablation for this benchmark is reported in Appendix~\ref{app:mv2vox_ablation}. Consistent with the super-resolution results, the full SDB objective gives the
strongest reconstruction fidelity and distributional similarity, indicating that the structural constraints remain complementary in the cross-modal multi-view-to-3D setting.

\section{Discussion}
\label{sec:conclusion}
We introduced \methodname \ as a constrained latent diffusion bridging framework that models modality translation as a heuristically constrained stochastic process. 
It induces structure into the bridge learning process, which  translates into additional inductive bias that restrict the solution space of valid bridges.
In contrast to methods that require paired data, this approach employs pairing as a data-driven heuristic within a broader diffusion-bridge framework.
As a by-product, by integrating this heuristic with structural constraints, the framework supports a range of supervision regimes. Experimental results indicate improved performance in fully paired settings, stable outcomes in semi-paired scenarios, and robust unpaired translation. Although modality translation remains under-constrained without full supervision, the findings suggest that latent diffusion bridges provide a principled framework
for composing inductive biases that restrict, but do not uniquely identify, admissible cross-modal couplings.

\paragraph{Limitations.}
Individual heuristics do not completely solve the ambiguity in modality translation. MM (Section~\ref{sec:marginal_matching}) matches the target marginal but may ignore the conditioning endpoint. Moreover, the WTA increases complexity as $K$ grows. CC (Section~\ref{sec:cycle_reversibility}) preserves content but does not ensure marginal matching. Using these methods together can help find bridge solutions with stronger couplings. Appendix~\ref{sec:heuristic_limitations} discusses these failure modes and their effects, and future work may introduce more constraints and inductive biases within the diffusion bridge framework.

\newpage

\section*{Impact Statement}

We introduced a method that tackles previously addressed tasks that are solved through Modality Translation. With this, we intend to drive the progress of machine learning, specifically in the context of Modality Translation through the application of methods that are commonly used recently, such as Denoising Diffusion Bridge Models and Latent Diffusion Models. Therefore, we neither expect nor intend to create a potential societal consequence that should be mentioned.

\clearpage

\bibliography{example_paper}
\bibliographystyle{icml2026}

\newpage
\appendix
\onecolumn

\section{Limitations and Complementarity of Heuristics}
\label{sec:heuristic_limitations}

Unpaired modality translation is non-identifiable in general: many joint distributions are consistent with the same endpoint marginals while inducing different conditional relationships between source and target variables. The heuristics considered in this work restrict the space of admissible couplings by imposing complementary structural constraints on the learned diffusion bridge. A qualitative summary is provided in Table~\ref{tab:heuristics_comparison}.

Endpoint marginal matching constrains the distribution of the target variable but leaves the conditional distribution given the source unconstrained. As a result, it does not restrict the dependence between source and target and admits solutions that ignore the conditioning endpoint. Cycle consistency introduces a reversibility constraint on the induced stochastic mapping, limiting many-to-one couplings that discard information about the source. However, enforcing consistency only at the endpoints does not constrain how probability mass is transported between them and can admit stochastic processes that satisfy endpoint reversibility while exhibiting incoherent or highly diffusive dynamics at intermediate times. Enforcing cycle consistency at intermediate diffusion times addresses this limitation by constraining the forward and reverse processes throughout the trajectory, thereby discouraging solutions that rely on endpoint-only cancellations and promoting coherent probability transport. Such endpoint-only solutions do not define stable conditional distributions, as their consistency depends on fragile cancellations rather than structured transport across time.

Paired supervision directly specifies a joint distribution and therefore resolves coupling ambiguity by construction, but requires access to paired data. Viewed jointly, marginal constraints restrict admissible endpoint distributions, reversibility-based constraints limit information loss from source to target, and paired supervision selects a specific coupling when available. Their complementary roles explain why combining multiple heuristics is necessary to obtain diffusion bridges that are both distributionally valid and conditionally informative, as summarized in Table~\ref{tab:heuristics_comparison}.

\begin{table*}[!h]
\centering
\small
\begin{tabular}{lcccc}
\toprule
\textbf{Heuristic} &
\textbf{Target marginal} &
\textbf{Coupling (uses condition)} &
\textbf{Flow coherence} &
\textbf{Exact correspondence} \\
\midrule
Marginal matching &
\checkmark &  &  &  \\
Cycle consistency (endpoint) &
 & \checkmark &  &  \\
Cycle consistency (trajectory) &
 & \checkmark & \checkmark &  \\
Data-driven (paired) &
\checkmark & \checkmark &  & \checkmark \\
\bottomrule
\end{tabular}
\caption{Qualitative comparison of heuristics and the failure modes they address. Marginal matching enforces validity under the target marginal but does not identify a coupling. Cycle constraints reduce coupling degeneracy by enforcing reversibility, with trajectory-level constraints additionally regularizing intermediate diffusion states. Paired supervision provides direct correspondence and can be composed with unpaired heuristics when available.}
\label{tab:heuristics_comparison}
\end{table*}

\section{Additional Training Details}
\label{app:training_details}

\paragraph{\textbf{Sampling and training procedure.}}
All diffusion bridges are trained and sampled using the standard iterative denoising scheme. More precisely, we use the iterative training procedure that alternates between optimizing the bridge and the autoencoders. At test time, translation starts by initializing the diffusion process with a sample from the target noise distribution and iteratively denoising it toward the target endpoint, conditioning on the source modality. We use the same discretization, noise schedule, and iterative update rules as in Latent Denoising Diffusion Bridge Models (LDDBM)~\cite{berman2025towards} to ensure a consistent and fair comparison.

\paragraph{\textbf{Number of diffusion steps.}}
We inherit the parameter $T=40$ denoising steps during inference from LDDBM~\cite{berman2025towards} in all experiments. The same number of steps applies to all supervision regimes, allowing for a fair comparison with previous works and across different supervision regimes.

\section{Synthetic Data Experiments}
\label{app:exp:synthetic_data}

\subsection{Data Generation}
\label{app:exp:synthetic_data_generation}

We construct a synthetic benchmark that enables controlled evaluation of diffusion bridges under unpaired, semi-paired, and fully paired supervision. The dataset is designed to exhibit nontrivial structure: while a ground-truth correspondence exists and is approximately reversible, the marginal distributions alone do not identify a unique coupling.

Each sample is generated from latent factors comprising a discrete \emph{content} variable, a discrete \emph{style} variable, and a continuous latent vector. Specifically, we sample a content label \(c \in \{1,\dots,K\}\), a style label \(s \in \{1,\dots,S\}\), and a continuous latent vector \(u \sim \mathcal{N}(0, I_d)\). Content-dependent mean vectors \(\mu^{(Y)}_c, \mu^{(X)}_c \in \mathbb{R}^d\) are fixed for each content class, while style-dependent linear transformations \(A_s, B_s \in \mathbb{R}^{d \times d}\) are sampled once and held fixed.

The conditioning endpoint \(z_0\) and target endpoint \(z_T\) are generated as
\[
\tilde z_0 = \mu^{(Y)}_c + A_s u + \epsilon_0, 
\qquad
\tilde z_T = \mu^{(X)}_c + B_s u + \epsilon_T,
\]
where \( \epsilon_0, \epsilon_T \sim \mathcal{N}(0, \sigma^2 I) \). The final endpoints are obtained by applying a fixed elementwise nonlinearity \( \phi(\cdot) \),
\[
z_0 = \phi(\tilde z_0), 
\qquad
z_T = \phi(\tilde z_T),
\]
which induces the marginal distributions \(p_{\mathcal{Y}}(z_0)\) and \(p_{\mathcal{X}}(z_T)\) used throughout the paper. In our experiments, we apply a fixed near-identity nonlinear warp
\[
\phi(z) = z + \alpha z^3,
\]
with a small coefficient \(\alpha>0\). This transformation preserves approximate invertibility while introducing sufficient nonlinearity to prevent the coupling from being recovered by linear transport alone. Without this nonlinear warp, the joint distribution remains Gaussian with a linear ground-truth coupling, making marginal matching largely sufficient and trivializing the unpaired translation task.

\begin{figure}[!h]
    \centering
    \includegraphics[width=\linewidth]{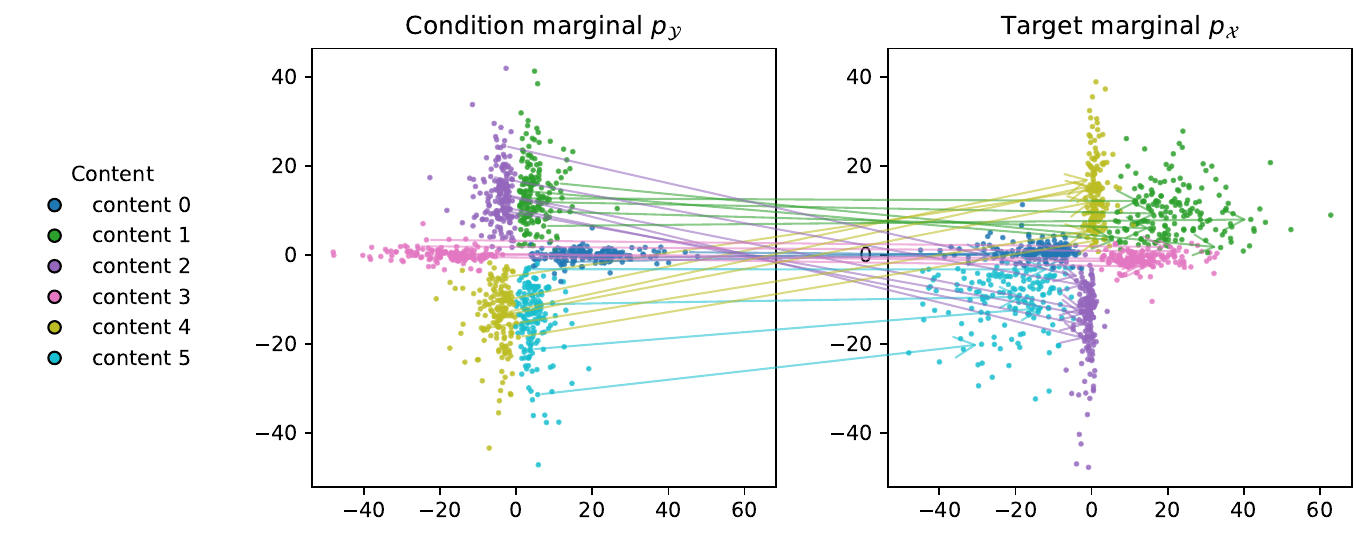}
    \caption{
    Synthetic 2D benchmark used in our experiments.
    \textbf{Left:} conditioning marginal \(p_{\mathcal{Y}}\).
    \textbf{Right:} target marginal \(p_{\mathcal{X}}\).
    Points are color-coded by the underlying content variable, which is shared across endpoints but not observed during training.
    Arrows indicate a small subset of true paired correspondences (shown for visualization only).
    While the two marginals appear broadly similar, the true coupling is highly ambiguous without additional structure, motivating the use of diffusion-bridge constraints beyond marginal matching.
    }
    \label{fig:synthetic_dataset}
\end{figure}

As illustrated in Figure~\ref{fig:synthetic_dataset}, this construction ensures that \(z_0\) and \(z_T\) share semantic content through the latent variable \(u\) and the content index \(c\), while style-dependent transformations introduce ambiguity in the marginal coupling. During training, supervision is controlled by selectively revealing paired samples \((z_0, z_T)\), allowing a systematic study of unpaired, semi-paired, and fully paired regimes.

\subsection{Denoiser Architecture}
\label{app:exp:syn_denoiser_architecture}

\begin{figure}[!h]
    \centering
    \includegraphics[width=\linewidth,trim=60 30 50 50,clip]{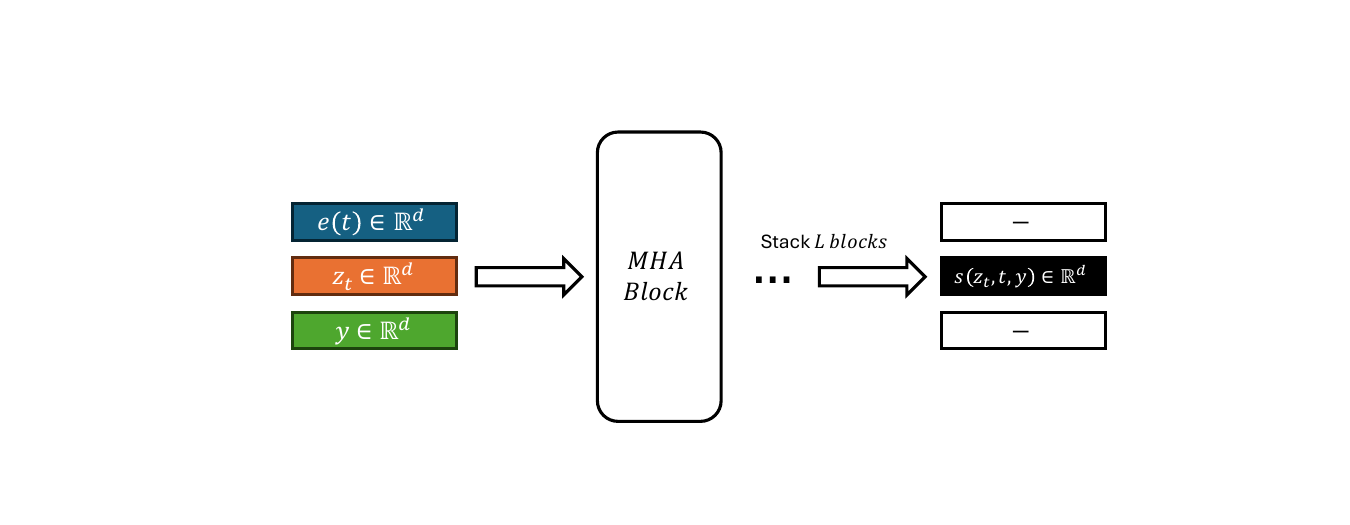}
    \caption{\textbf{Denoiser architecture for the synthetic benchmark.}
The noisy latent \(z_t\), conditioning endpoint \(y\), and diffusion-time embedding \(e(t)\) are represented as a set of tokens of dimensionality \(d\) and processed jointly by a stack of \(L\) multi-head attention blocks. The output corresponding to the \(z_t\) token is used directly as the score (or noise) prediction for denoising.
}
    \label{fig:syn_denoiser_architecture}
\end{figure}

As depicted in Figure~\ref{fig:syn_denoiser_architecture}, the denoiser is implemented as a stack of \(L\) multi-head attention (MHA) blocks. The network operates directly in the endpoint space, i.e., it implements a denoising diffusion bridge model \cite{zhou2024denoising} without modality-specific encoders and decoders, as opposed to its latent variant \cite{berman2025towards}, and is conditioned on the source endpoint and diffusion time. At each diffusion step, the noisy latent vector \(z_t \in \mathbb{R}^d\), the conditioning endpoint \(y \in \mathbb{R}^d\), and a diffusion time embedding \(e(t) \in \mathbb{R}^{d}\) (implemented using standard sinusoidal embeddings) are represented as a set of tokens, each of dimensionality \(d\), and processed jointly by the attention blocks. The token corresponding to \(z_t\) serves as the prediction target, and the output of the final attention block at this token is used directly as the denoiser output. This output is interpreted as a score (or noise) prediction according to the diffusion parameterization used in the main text.

This simple design is sufficient for the synthetic benchmark and intentionally avoids architectural complexity, allowing us to isolate the effect of diffusion bridge constraints and supervision regimes.

\subsection{Evaluation Metrics}
\label{sec:synthetic_metrics}

We evaluate synthetic translation performance using the following metrics, which separately assess marginal alignment, conditional coupling quality, and reversibility.

\begin{itemize}
    \item \textbf{Target marginal alignment (SWD).}
    We measure sliced Wasserstein distance (SWD) between generated target samples \(\hat{x} \sim p_\theta(x \mid y)\) and samples \(x \sim p_{\mathcal{X}}\). SWD is computed by projecting samples onto random one-dimensional directions and averaging the Wasserstein-1 distance across projections. Lower values indicate better agreement with the target marginal.

    \item \textbf{Target marginal alignment (MMD).}
    We additionally report the maximum mean discrepancy (MMD) between generated samples \(\{\hat{x}_i\}_{i=1}^N\) and target samples \(\{x_j\}_{j=1}^M\), defined as
    \[
    \mathrm{MMD}^2 = 
    \mathbb{E}_{x,x'}[k(x,x')] 
    + \mathbb{E}_{\hat{x},\hat{x}'}[k(\hat{x},\hat{x}')] 
    - 2\,\mathbb{E}_{x,\hat{x}}[k(x,\hat{x})],
    \]
    where \(k(\cdot,\cdot)\) is an RBF kernel. In practice we estimate \(\mathrm{MMD}^2\) using the unbiased U-statistic estimator on finite samples; lower values indicate closer alignment of the target marginal distributions.

    \item \textbf{Conditional content accuracy.}
    To evaluate coupling quality, we train a classifier \(h:\mathbb{R}^d \rightarrow \{1,\dots,K\}\) on target samples with known content labels and report the accuracy of \(h(\hat{x})\) on translated samples \(\hat{x}\). This metric measures whether conditioning on the source endpoint preserves the underlying semantic content.

    \item \textbf{Endpoint cycle error.}
    Reversibility is quantified by generating a forward translation \(\hat{x} \sim p_\theta(x \mid y)\) followed by a reverse translation \(\tilde{y} \sim p_\theta(y \mid \hat{x})\), and reporting the mean squared error \(\mathbb{E}\|\tilde{y} - y\|_2^2\). Lower values indicate improved cycle consistency.

    Endpoint cycle error serves as a diagnostic for coupling consistency, penalizing solutions that achieve good marginal alignment and content preservation yet discard information about the specific conditioning endpoint.

\end{itemize}

\subsection{Additional Experiments}
\label{app:exp:synthetic_data_results}

\begin{figure}
\begin{tikzpicture}

\definecolor{KsixColor}{HTML}{1F77B4}
\definecolor{KtwentyColor}{HTML}{FF7F0E}
\definecolor{KhundredColor}{HTML}{2CA02C}
\definecolor{KthousandColor}{HTML}{D62728}

\pgfplotstableread[col sep=space]{
rho pairedSwd oursSwd pairedMmd oursMmd pairedAcc oursAcc pairedCycle oursCycle
0 0.021336 0.019676 -0.000057 -0.000111 0.183000 0.868333 0.977830 0.679605
0.1 0.022075 0.021102 -0.000037 -0.000093 0.355000 0.883667 0.957288 0.668844
0.5 0.019514 0.019300 -0.000133 -0.000151 0.641000 0.954667 0.844565 0.615000
1 0.023158 0.019372 -0.000019 -0.000147 0.887333 0.965000 0.685644 0.603987
}\dataKsix

\pgfplotstableread[col sep=space]{
rho pairedSwd oursSwd pairedMmd oursMmd pairedAcc oursAcc pairedCycle oursCycle
0 0.020907 0.023040 -0.000111 -0.000033 0.051333 0.860000 1.016587 0.676029
0.1 0.019956 0.019988 -0.000166 -0.000154 0.267333 0.882667 0.996782 0.658263
0.5 0.021758 0.021029 -0.000041 -0.000131 0.590000 0.937667 0.875019 0.606545
1 0.019344 0.018911 -0.000176 -0.000169 0.870333 0.956333 0.682346 0.597807
}\dataKtwenty

\pgfplotstableread[col sep=space]{
rho pairedSwd oursSwd pairedMmd oursMmd pairedAcc oursAcc pairedCycle oursCycle
0 0.022169 0.019815 -0.000085 -0.000148 0.012667 0.823000 1.038560 0.676324
0.1 0.020743 0.021101 -0.000107 -0.000115 0.212333 0.856000 1.014405 0.655008
0.5 0.021579 0.020361 -0.000083 -0.000146 0.575000 0.919667 0.882816 0.604534
1 0.020383 0.019984 -0.000124 -0.000192 0.844667 0.924667 0.682794 0.596735
}\dataKhundred

\pgfplotstableread[col sep=space]{
rho pairedSwd oursSwd pairedMmd oursMmd pairedAcc oursAcc pairedCycle oursCycle
0 0.023775 0.020934 0.000043 -0.000122 0.001333 0.632000 1.046574 0.685281
0.1 0.024824 0.021518 0.000027 -0.000098 0.155667 0.643333 1.023749 0.661377
0.5 0.021264 0.020984 -0.000113 -0.000144 0.420000 0.697333 0.883517 0.610445
1 0.021864 0.020858 -0.000097 -0.000149 0.649000 0.703000 0.678674 0.596766
}\dataKthousand

\newcommand{\PlotPairT}[4]{%
  \addplot[dashed,color=#1] table[x=rho,y={#3}]{#2};
  \addplot[solid,color=#1]  table[x=rho,y={#4}]{#2};
}

\newcommand{\PlotPairTDropZeroPaired}[4]{%
\addplot[dashed,color=#1, dropRhoZero] table[x=rho,y={#3}]{#2};
\addplot[solid,color=#1] table[x=rho,y={#4}]{#2};
}

\pgfplotsset{
dropRhoZero/.style={
x filter/.code={
\pgfmathparse{\pgfmathresult==0 ? nan : \pgfmathresult}%
}
}
}

\begin{groupplot}[
  group style={group size=2 by 2, horizontal sep=2cm, vertical sep=2.0cm},
  width=0.4\textwidth, height=4.5cm, scale only axis, grid=both,
  xlabel={$\rho$}, xmin=0, xmax=1,
  tick label style={font=\scriptsize}, every axis plot/.append style={line width=0.9pt},
]

\nextgroupplot[title={Coupling quality}, ylabel={Content acc.\ $\rightarrow$}, ymin=0.0, ymax=1.0,
  ]
\addplot[dashed,color=KsixColor,forget plot] table[x=rho,y=pairedAcc, dropRhoZero]{\dataKsix};
\addplot[solid,color=KsixColor,forget plot]  table[x=rho,y=oursAcc]{\dataKsix};
\addplot[dashed,color=KtwentyColor,forget plot] table[x=rho,y=pairedAcc, dropRhoZero]{\dataKtwenty};
\addplot[solid,color=KtwentyColor,forget plot]  table[x=rho,y=oursAcc]{\dataKtwenty};
\addplot[dashed,color=KhundredColor,forget plot] table[x=rho,y=pairedAcc, dropRhoZero]{\dataKhundred};
\addplot[solid,color=KhundredColor,forget plot]  table[x=rho,y=oursAcc]{\dataKhundred};
\addplot[dashed,color=KthousandColor,forget plot] table[x=rho,y=pairedAcc, dropRhoZero]{\dataKthousand};
\addplot[solid,color=KthousandColor,forget plot]  table[x=rho,y=oursAcc]{\dataKthousand};

\nextgroupplot[title={Reversibility}, ylabel={Cycle MSE $\leftarrow$},]
\PlotPairTDropZeroPaired{KsixColor}{\dataKsix}{pairedCycle}{oursCycle}
\PlotPairTDropZeroPaired{KtwentyColor}{\dataKtwenty}{pairedCycle}{oursCycle}
\PlotPairTDropZeroPaired{KhundredColor}{\dataKhundred}{pairedCycle}{oursCycle}
\PlotPairTDropZeroPaired{KthousandColor}{\dataKthousand}{pairedCycle}{oursCycle}

\nextgroupplot[title={Marginal alignment (SWD)}, ylabel={SWD $\leftarrow$},]
\PlotPairTDropZeroPaired{KsixColor}{\dataKsix}{pairedSwd}{oursSwd}
\PlotPairTDropZeroPaired{KtwentyColor}{\dataKtwenty}{pairedSwd}{oursSwd}
\PlotPairTDropZeroPaired{KhundredColor}{\dataKhundred}{pairedSwd}{oursSwd}
\PlotPairTDropZeroPaired{KthousandColor}{\dataKthousand}{pairedSwd}{oursSwd}

\nextgroupplot[title={Marginal alignment (MMD$^2$)}, ylabel={MMD$^2$ $\leftarrow$},]
\PlotPairTDropZeroPaired{KsixColor}{\dataKsix}{pairedMmd}{oursMmd}
\PlotPairTDropZeroPaired{KtwentyColor}{\dataKtwenty}{pairedMmd}{oursMmd}
\PlotPairTDropZeroPaired{KhundredColor}{\dataKhundred}{pairedMmd}{oursMmd}
\PlotPairTDropZeroPaired{KthousandColor}{\dataKthousand}{pairedMmd}{oursMmd}

\end{groupplot}


\begin{axis}[
hide axis,
width=3cm,
height=2cm,
legend to name=appendixLegend,
legend columns=4,
legend style={
/tikz/column sep=5pt,
draw=none,
font=\normalsize,
nodes={align=center},
legend cell align=center,
},
]
\addplot[solid,draw=none,color=KsixColor] coordinates {(0,0) (1,0)};
\addlegendentry{$K=6$}
\addplot[solid,draw=none,color=KtwentyColor] coordinates {(0,0) (1,0)};
\addlegendentry{$K=20$}
\addplot[solid,draw=none,color=KhundredColor] coordinates {(0,0) (1,0)};
\addlegendentry{$K=100$}
\addplot[solid,draw=none,color=KthousandColor] coordinates {(0,0) (1,0)};
\addlegendentry{$K=1000$}
\end{axis}

\begin{axis}[
hide axis,
width=3cm,
height=2cm,
legend to name=appendixLegendMethod,
legend columns=2,
legend style={/tikz/column sep=5pt, nodes={inner xsep=6pt, inner ysep=3pt}, draw=none, font=\normalsize, nodes={align=center}, legend cell align=center},
]
\addplot[black,solid,draw=none]  coordinates {(0,0) (1,0)};
\addlegendentry{Semi-paired (ours)}
\addplot[black,dashed,draw=none] coordinates {(0,0) (1,0)};
\addlegendentry{Paired-only}
\end{axis}

\path (current bounding box.south) node[
anchor=north,
yshift=-10pt,
draw=black,
rounded corners=5pt,
fill=white,
line width=0.4pt,
inner sep=2pt,
] {%
\begin{tabular}{c}
\pgfplotslegendfromname{appendixLegend}\\[-1pt]
\pgfplotslegendfromname{appendixLegendMethod}
\end{tabular}%
};
\end{tikzpicture}

\caption{\textbf{Synthetic benchmark results across supervision levels (\(\rho\)) for varying \(K\)}. As the number of modes increases (\(K\in\{6,20,100,1000\}\)), semi-paired training maintains strong coupling quality and cycle consistency relative to paired-only baselines, while preserving competitive target marginal alignment.}
\label{fig:synthetic_rho_grid_allK}
\end{figure}
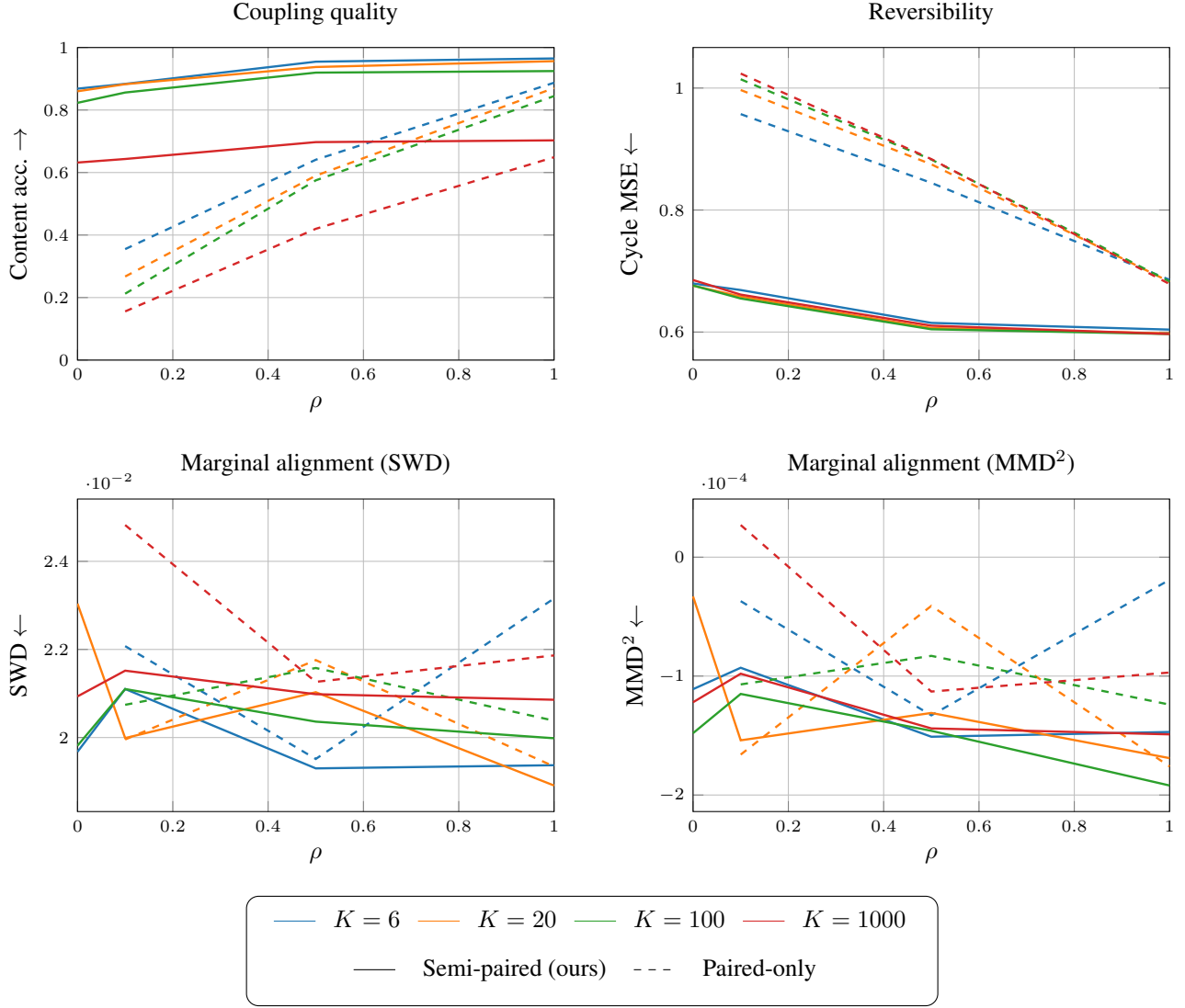

We extend the synthetic study from the main text (reported for $K{=}6$) by increasing the complexity of the problem. This is done by increasing the number of content labels, i.e.\ the parameter $K$ over $\{6,20,100,1000\}$ while varying the supervision level $\rho \in [0,1]$. The experimental protocol, objectives, and evaluation metrics are unchanged: we report coupling quality via content accuracy, reversibility via endpoint cycle MSE, and marginal alignment via SWD and MMD$^2$. We compare our semi-paired training procedure against a paired-only baseline trained with the same amount of pairing signal. Figure~\ref{fig:synthetic_rho_grid_allK} summarizes the results for the whole sweep.

Across all $K$, the qualitative trends observed in the $K{=}6$ setting persist. In the fully unpaired regime ($\rho=0$), only marginal-matching and cycle-consistency are applicable. In this setting, training with our structural constraints remains robust as $K$ grows, achieving high coupling quality and low cycle error even without explicit pairing: content accuracy stays above $0.82$ for $K\le 100$ and remains substantially high even at $K{=}1000$ ($0.632$), while cycle MSE is consistently reduced to $\approx 0.68$ across all $K$.

For any $\rho>0$, we can compare directly to the paired-only baseline. At low supervision (e.g., $\rho=0.1$), the paired-only baseline increasingly struggles to learn condition-informative couplings as $K$ grows, with content accuracy dropping from $0.183$ at $K{=}6$ to near-zero at larger $K$, and cycle MSE remaining high. In contrast, semi-paired training yields substantially better coupling quality and reversibility at the same supervision level, indicating that the structural constraints reduce ambiguity that pairing alone does not address in the low-$\rho$ regime.

As supervision increases, both methods improve, but the relative advantage of semi-paired training is preserved throughout the sweep. At intermediate supervision (e.g., $\rho=0.5$), our method exhibits uniformly higher content accuracy and lower cycle MSE across all $K$, suggesting that the added structural regularization continues to provide complementary signal beyond limited pairing. Even in the fully paired regime ($\rho=1$), semi-paired training remains complementary to full supervision: for every $K$ we observe a consistent gap in favor of our method, alongside a nontrivial reduction in cycle MSE.

Finally, marginal alignment metrics (SWD and MMD$^2$) remain stable across $K$ and across $\rho$, with only small fluctuations. This reinforces the interpretation that the gains in coupling quality and reversibility are not explained by better marginal matching, but by improved structural alignment of the learned stochastic coupling. Overall, the $K$-sweep supports the conclusion that as the synthetic task becomes more multimodal and under-constrained, explicit structural heuristics become increasingly important for identifying semantically meaningful and reversible couplings, and these benefits persist even when pairing supervision is available.

\subsection{WTA Assignment Sensitivity and Computational Cost}
\label{app:capacity_sensitivity}
We evaluate the sensitivity of the WTA assignment to two implementation choices:
the per-condition capacity $C_y$ and the number of candidate endpoints $K$.
Table~\ref{tab:synthetic_capacity} shows that the capacity constraint is beneficial
but not brittle: moderate capacities improve coupling quality by preventing
condition dominance, while very large capacities approach the unconstrained
assignment and reduce performance. Table~\ref{tab:wta_runtime_synth} shows that
increasing $K$ improves content accuracy by allowing better candidate matches,
but the gains saturate while runtime grows approximately linearly.
\begin{table}[!h]
\centering
\small
\setlength{\tabcolsep}{8pt}
\renewcommand{\arraystretch}{1.1}
\begin{tabular}{lcc}
\toprule
Capacity constraint $C_y$ & Content accuracy $\uparrow$ & Gain over $C_{\infty}$ \\
\midrule
1   & 0.941 & +0.020 \\
2   & \textbf{0.955} & \textbf{+0.034} \\
4   & 0.954 & +0.033 \\
8   & 0.949 & +0.028 \\
16  & 0.937 & +0.014 \\
$\infty$ & 0.921 & 0.000 \\
\bottomrule
\end{tabular}
\caption{Effect of the WTA capacity constraint on content accuracy in the synthetic benchmark. We use $C_{\infty}$ as the unconstrained reference. Moderate capacities yield the best coupling quality, with $C_y=2$--$4$ achieving the highest content accuracy, while the unconstrained regime is slightly worse.}
\label{tab:synthetic_capacity}
\end{table}

\begin{table}[!h]
\centering
\small
\setlength{\tabcolsep}{5.5pt}
\renewcommand{\arraystretch}{1.12}
\begin{tabular}{c c c c c}
\toprule
WTA candidates $K$ & Train runtime / iter (s) $\downarrow$ & Relative runtime ($\times$) $\downarrow$ & Content accuracy $\uparrow$ & Gain over $K{=}1$ $\uparrow$ \\
\midrule
1  & 0.42 & 1.00  & 0.845 & 0.000 \\
2  & 0.68 & 1.62  & 0.899 & 0.054 \\
4  & 1.19 & 2.83  & 0.931 & 0.086 \\
8  & 2.21 & 5.26  & 0.955 & 0.12 \\
16 & 4.25 & 10.12 & 0.971 & 0.126 \\
\bottomrule
\end{tabular}
\caption{\textbf{Training-time overhead of WTA on the synthetic setup.} We report per-iteration training runtime and content accuracy as a function of the number of WTA candidates $K$. In our implementation, WTA candidates are processed serially, so increasing $K$ primarily increases runtime through repeated candidate evaluation linearly, while content accuracy improves with diminishing returns. We see that accuracy saturates after $K=8$.}
\label{tab:wta_runtime_synth}
\end{table}

\newpage
\section{Super-Resolution Experimental Details}
\label{app:exp:sr}

\subsection{Architectures}
\label{app:sr_arch}

\paragraph{\textbf{Score model.}}
We use the same denoiser (score model) DTDiT architecture as in LDDBM \cite{berman2025towards} for the conditional bridge \(s_\theta(z_t,t\mid z_T)\), including the same conditioning mechanism on the LR endpoint and the same diffusion-time embedding. For fair comparison and to isolate the effect of the proposed method, unless stated otherwise, we match the LDDBM training budget and optimization settings for this benchmark.

\paragraph{\textbf{Autoencoders.}}
Our bridge model assumes that the LR and HR endpoints admit a sufficiently expressive and consistent shared latent space. Accordingly, we adopt the same encoder/decoder family as LDDBM \cite{berman2025towards} and use the pretrained autoencoder from latent diffusion models (LDM) \cite{rombach2021highresolution} for the SR benchmark.

\subsection{Evaluation Protocol}
\label{sec:sr_eval_protocol}

To demonstrate the superiority of \methodname, we follow the evaluation protocol from previous works \cite{moser2024waving, berman2025towards}. The model is trained on Flickr-Faces-HQ (FFHQ) \cite{karras2019style} and evaluated zero-shot on CelebA-HQ \cite{karras2017progressive}. The task is defined as translation from resolution of \(16 \times 16\) to \(128 \times 128\). We compare \methodname~ to the data-driven bridge baseline \cite{berman2025towards} used in the main text and to the task-specific diffusion-based baseline DiWa \cite{moser2024waving}.

\subsection{Metrics}
\label{sec:sr_metrics}

We evaluate SR performance using the following metrics. PSNR emphasizes low-level reconstruction accuracy and penalizes pixel-wise deviations, SSIM captures preservation of local structure (luminance/contrast/edges) that is critical for maintaining image content, and LPIPS measures similarity in deep feature space, reflecting perceptual quality and realism that may not be captured by pixel-level criteria. Let \(x \in [0,1]^{H \times W \times C}\) denote the ground-truth HR image and \(\hat{x} \in [0,1]^{H \times W \times C}\) the generated HR image (with \(H=W=128\) in our setting).

\begin{itemize}
    \item \textbf{Pixel-level fidelity (PSNR).}
    We report peak signal-to-noise ratio (PSNR) between \(\hat{x}\) and \(x\), defined via the mean squared error
    \[
    \mathrm{MSE}(x,\hat{x}) \;=\; \frac{1}{HWC}\sum_{h=1}^{H}\sum_{w=1}^{W}\sum_{c=1}^{C}\big(x_{h,w,c}-\hat{x}_{h,w,c}\big)^2,
    \]
    and
    \[
    \mathrm{PSNR}(x,\hat{x}) \;=\; 10 \log_{10}\!\left(\frac{\mathrm{MAX}^2}{\mathrm{MSE}(x,\hat{x})}\right),
    \]
    where \(\mathrm{MAX}\) is the maximum possible pixel value (here \(\mathrm{MAX}=1\)). Higher PSNR indicates better pixel-level fidelity.

    \item \textbf{Structural similarity (SSIM).}
    We report structural similarity index (SSIM) between \(\hat{y}\) and \(y\). For image patches with means \(\mu_y,\mu_{\hat{y}}\), variances \(\sigma_y^2,\sigma_{\hat{y}}^2\), and covariance \(\sigma_{y\hat{y}}\), SSIM is
    \[
    \mathrm{SSIM}(x,\hat{x}) \;=\;
    \frac{(2\mu_x\mu_{\hat{x}}+C_1)(2\sigma_{x\hat{x}}+C_2)}
         {(\mu_x^2+\mu_{\hat{x}}^2+C_1)(\sigma_x^2+\sigma_{\hat{x}}^2+C_2)},
    \]
    where \(C_1=(K_1\mathrm{MAX})^2\) and \(C_2=(K_2\mathrm{MAX})^2\) are stabilization constants (with standard choices \(K_1=0.01\), \(K_2=0.03\)). We use the standard windowing/averaging procedure over the image as in \cite{moser2024waving}. Higher SSIM indicates better structural preservation.

    \item \textbf{Perceptual similarity (LPIPS).}
    We report learned perceptual image patch similarity (LPIPS) between \(\hat{y}\) and \(y\), computed as a distance between deep feature activations of a pretrained network. Let \(\phi_l(\cdot)\) denote activations at layer \(l\), spatially normalized, and \(w_l\) learned linear weights. LPIPS is
    \[
    \mathrm{LPIPS}(x,\hat{x}) \;=\; \sum_{l \in \mathcal{L}} \frac{1}{H_lW_l}\sum_{h=1}^{H_l}\sum_{w=1}^{W_l}
    \left\|\, w_l \odot \big(\phi_l(x)_{h,w} - \phi_l(\hat{x})_{h,w}\big)\right\|_2^2,
    \]
    where \(\odot\) denotes elementwise multiplication and \((H_l,W_l)\) is the spatial resolution at layer \(l\). Lower LPIPS indicates higher perceptual similarity and typically better perceptual realism.
\end{itemize}

\newpage
\subsection{Additional Results}
\subsubsection{Real-data ablation of structural constraints}
\label{app:sr_ablation}
To complement the synthetic ablation in Table~\ref{tab:heuristic_ablation}, we ablate the same components on SR in Table~\ref{tab:sr_ablation}. Marginal matching alone improves target-marginal validity but is insufficient to preserve source-conditioned structure. Endpoint cycle consistency improves reconstruction fidelity by discouraging information loss, while trajectory cycle consistency gives the strongest results by regularizing the intermediate bridge path. The trend is consistent with the synthetic benchmark: MM primarily supports marginal validity, endpoint CC improves reversibility, and trajectory CC strengthens flow coherence.
\begin{table*}[h]
\centering
\caption{\textbf{Super-Resolution Performance across Supervision Regimes.} 
Evaluation of image reconstruction quality and distributional alignment. PSNR and SSIM measure pixel-wise and structural fidelity, while LPIPS assesses perceptual consistency.}
\label{tab:sr_ablation}
\begin{tabular}{llccc}
\toprule
$\rho$ & Method
& PSNR $\uparrow$
& SSIM $\uparrow$
& LPIPS $\downarrow$ \\
\midrule
\multirow{3}{*}{0}
& Marginal matching only
& $17.8{\pm}0.7$ & $0.49{\pm}0.05$ & $0.42{\pm}0.04$ \\
& + Endpoint cycle
& $18.5{\pm}0.6$ & $0.52{\pm}0.05$ & $0.39{\pm}0.04$ \\
& \textbf{+ Trajectory cycle}
& $\mathbf{19.0{\pm}0.6}$ & $\mathbf{0.54{\pm}0.05}$ & $\mathbf{0.37{\pm}0.04}$ \\
\midrule
\multirow{5}{*}{0.5}
& Marginal matching only
& $17.8{\pm}0.7$ & $0.49{\pm}0.05$ & $0.42{\pm}0.04$ \\
& + Endpoint cycle
& $18.5{\pm}0.6$ & $0.52{\pm}0.05$ & $0.39{\pm}0.04$ \\
& + Trajectory cycle
& $19.0{\pm}0.6$ & $0.54{\pm}0.05$ & $0.37{\pm}0.04$ \\
& Paired-only
& $24.9{\pm}0.3$ & $0.67{\pm}0.02$ & $0.33{\pm}0.01$ \\
& \textbf{Semi-paired (ours)}
& $\mathbf{25.2{\pm}0.3}$ & $\mathbf{0.68{\pm}0.02}$ & $\mathbf{0.32{\pm}0.01}$ \\
\midrule
\multirow{5}{*}{1.0}
& Marginal matching only
& $17.8{\pm}0.7$ & $0.49{\pm}0.05$ & $0.42{\pm}0.04$ \\
& + Endpoint cycle
& $18.5{\pm}0.6$ & $0.52{\pm}0.05$ & $0.39{\pm}0.04$ \\
& + Trajectory cycle
& $19.0{\pm}0.6$ & $0.54{\pm}0.05$ & $0.37{\pm}0.04$ \\
& Paired-only
& $25.6{\pm}0.4$ & $0.68{\pm}0.03$ & $0.32{\pm}0.01$ \\
& \textbf{Semi-paired (ours)}
& $\mathbf{25.9{\pm}0.3}$ & $\mathbf{0.69{\pm}0.02}$ & $\mathbf{0.31{\pm}0.01}$ \\
\bottomrule
\end{tabular}
\end{table*}

\subsubsection{Finer Supervision Sweep}
\label{app:sr_finer_sweep}

We additionally evaluate intermediate paired fractions beyond the main
$\rho \in \{0,0.5,1.0\}$ setting. The total number of endpoint samples is fixed
while only the fraction of paired correspondences changes. The results show a
gradual degradation as paired supervision is reduced, supporting the claim that
SDB benefits from structural constraints across supervision levels rather than
only at the three coarse settings reported in the main text. As shown in Table~\ref{tab:sr_main_expanded_copy}, performance changes smoothly across
intermediate paired fractions, indicating that the main-text results are not an
artifact of evaluating only $\rho \in \{0,0.5,1.0\}$.

\begin{table*}[!h]
\centering
\footnotesize
\setlength{\tabcolsep}{3pt}
\renewcommand{\arraystretch}{1.15}
\caption{Zero-shot SR on CelebA-HQ when training on FFHQ, sweeping paired fraction $\rho$.}
\label{tab:sr_main_expanded_copy}
\resizebox{\textwidth}{!}{%
\begin{tabular}{l c ccc ccc ccc ccc}
\toprule
& \multicolumn{1}{c}{$\rho=0$}
& \multicolumn{3}{c}{$\rho=0.25$}
& \multicolumn{3}{c}{$\rho=0.5$}
& \multicolumn{3}{c}{$\rho=0.75$}
& \multicolumn{3}{c}{$\rho=1.0$} \\
\cmidrule(lr){2-2} \cmidrule(lr){3-5} \cmidrule(lr){6-8} \cmidrule(lr){9-11} \cmidrule(lr){12-14}
& SDB
& DiWa & LDDBM & SDB
& DiWa & LDDBM & SDB
& DiWa & LDDBM & SDB
& DiWa & LDDBM & SDB \\
\midrule
PSNR$\uparrow$
& $19.0{\pm}0.6$
& $20.7{\pm}0.4$ & $22.4{\pm}0.4$ & $24.0{\pm}0.4$
& $22.6{\pm}0.2$ & $24.9{\pm}0.3$ & $25.2{\pm}0.3$
& $23.1{\pm}0.2$ & $25.4{\pm}0.3$ & $25.6{\pm}0.3$
& $23.3{\pm}0.2$ & $25.6{\pm}0.4$ & $25.9{\pm}0.3$ \\
SSIM$\uparrow$
& $0.54{\pm}0.05$
& $0.56{\pm}0.03$ & $0.62{\pm}0.03$ & $0.64{\pm}0.03$
& $0.62{\pm}0.02$ & $0.67{\pm}0.02$ & $0.68{\pm}0.02$
& $0.64{\pm}0.02$ & $0.674{\pm}0.03$ & $0.686{\pm}0.02$
& $0.65{\pm}0.02$ & $0.68{\pm}0.03$ & $0.69{\pm}0.02$ \\
LPIPS$\downarrow$
& $0.37{\pm}0.04$
& $0.43{\pm}0.02$ & $0.36{\pm}0.02$ & $0.34{\pm}0.02$
& $0.40{\pm}0.01$ & $0.33{\pm}0.01$ & $0.32{\pm}0.01$
& $0.39{\pm}0.01$ & $0.326{\pm}0.01$ & $0.314{\pm}0.01$
& $0.39{\pm}0.01$ & $0.32{\pm}0.01$ & $0.31{\pm}0.01$ \\
\bottomrule
\end{tabular}}
\end{table*}

\newpage
\subsubsection{Qualitative Results}
\label{app:sr_results}
\begin{figure}[h!]
\centering
\setlength{\tabcolsep}{10pt}
\renewcommand{\arraystretch}{1.15}

\begin{tabular}{>{\raggedleft\arraybackslash}m{1.8cm} *{3}{m{0.22\textwidth}}}

\textbf{Original} 
& \includegraphics[width=\linewidth]{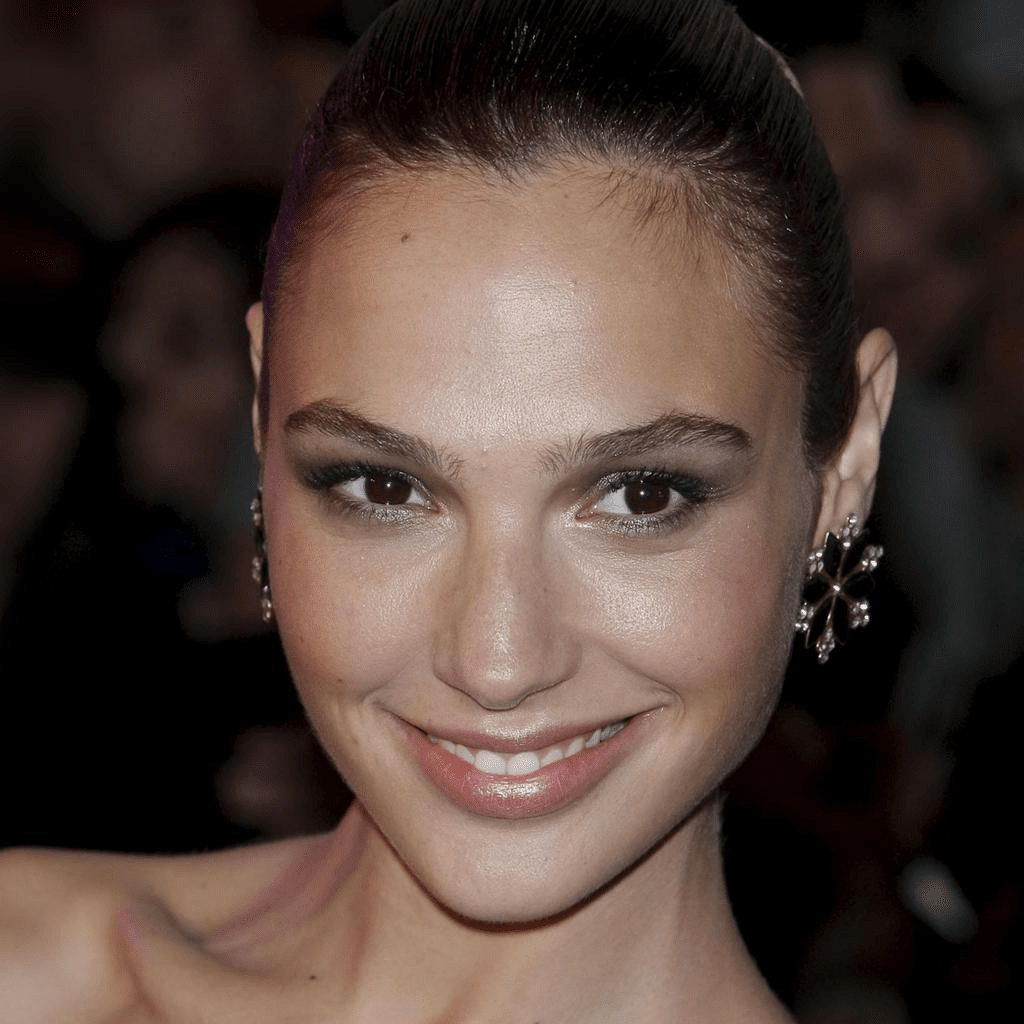} 
& \includegraphics[width=\linewidth]{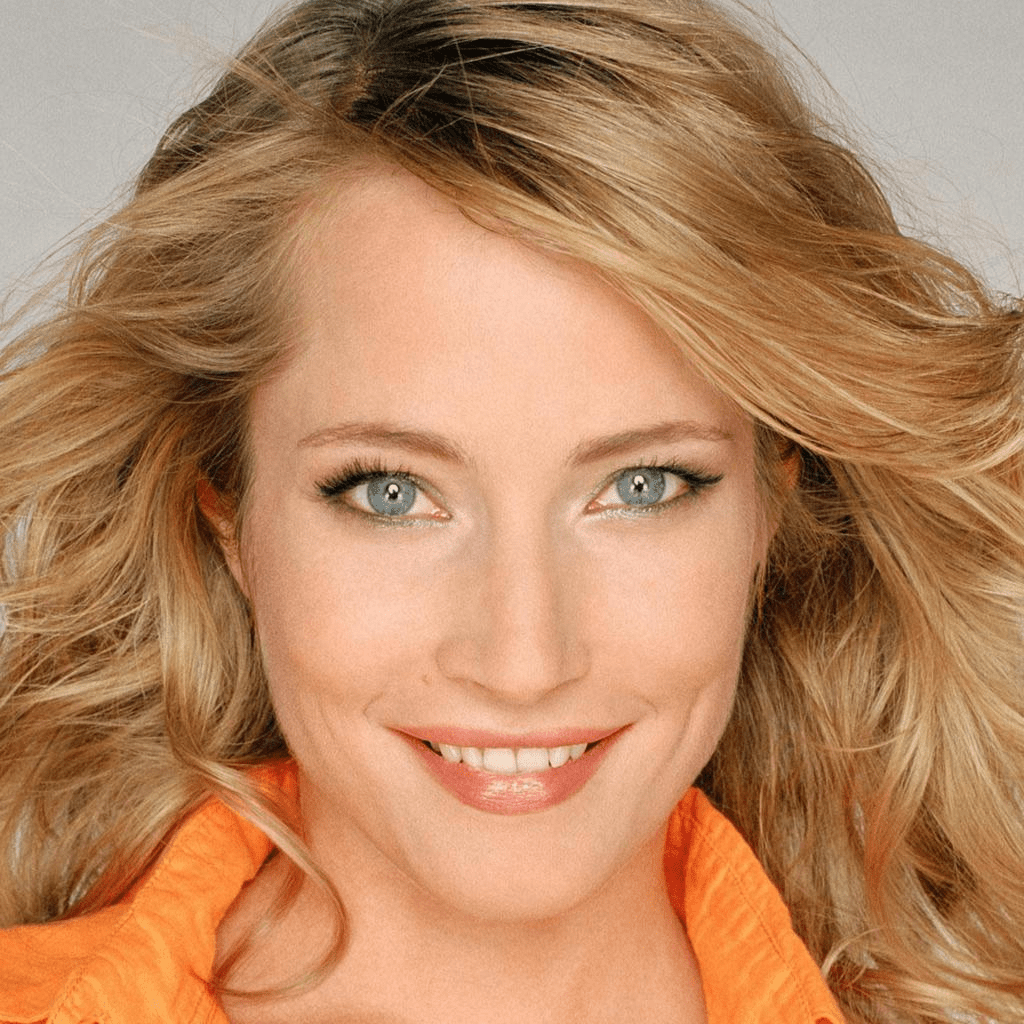} 
& \includegraphics[width=\linewidth]{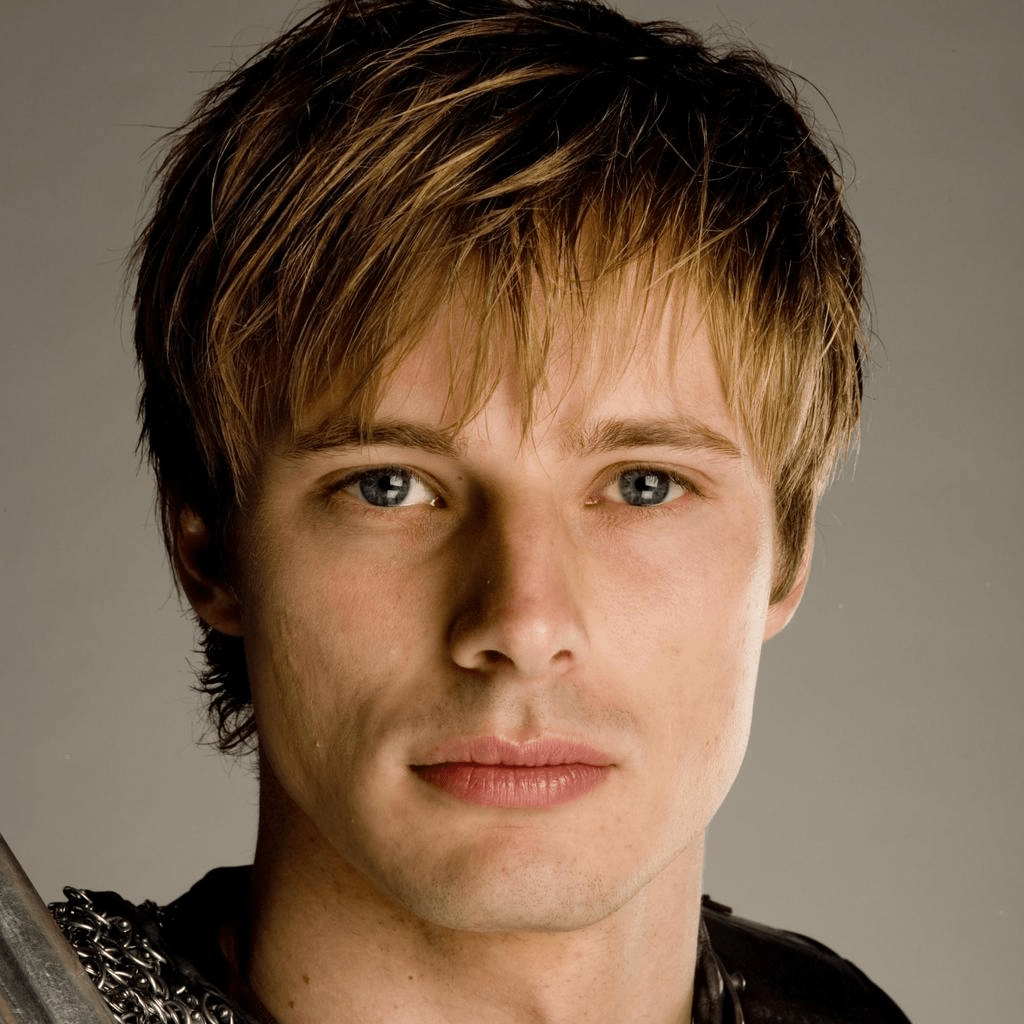} \\

\textbf{Down sampled} 
& \includegraphics[width=\linewidth]{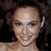} 
& \includegraphics[width=\linewidth]{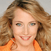} 
& \includegraphics[width=\linewidth]{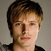} \\

\textbf{DiWa} 
& \includegraphics[width=\linewidth]{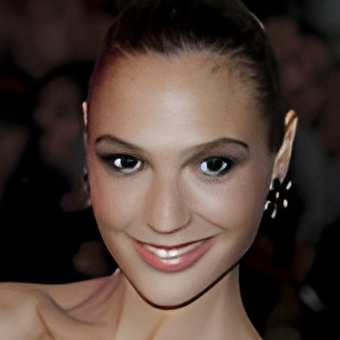} 
& \includegraphics[width=\linewidth]{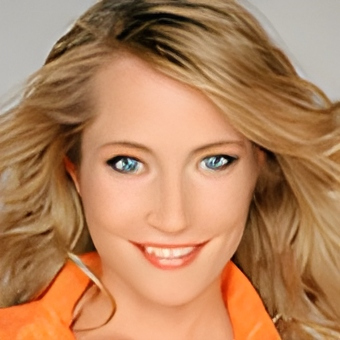} 
& \includegraphics[width=\linewidth]{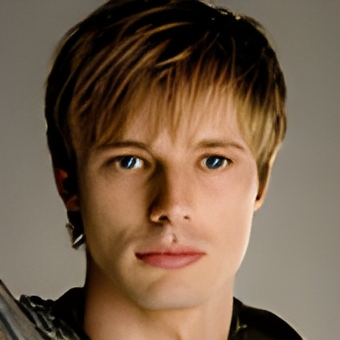} \\

\textbf{LDDBM} 

& \includegraphics[width=\linewidth]{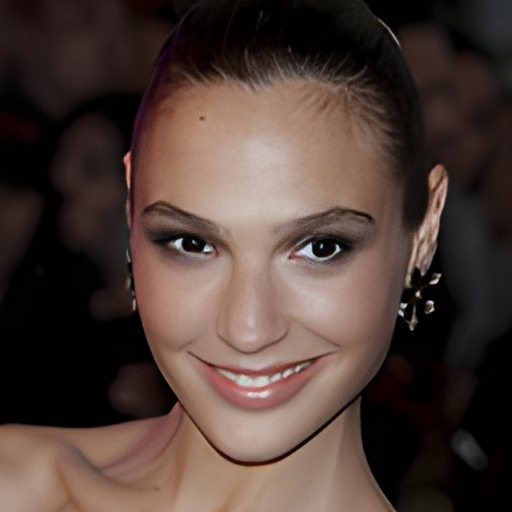} 
& \includegraphics[width=\linewidth]{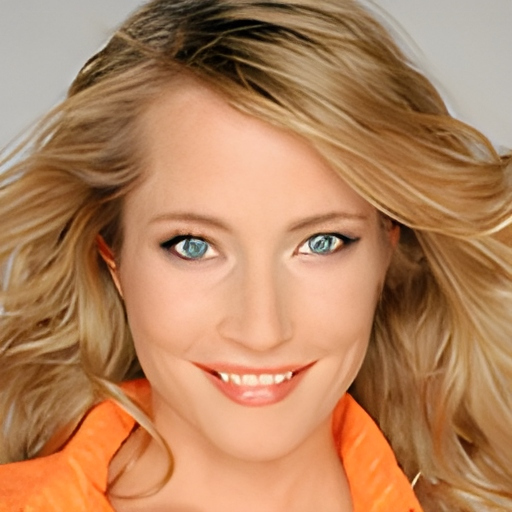} 
& \includegraphics[width=\linewidth]{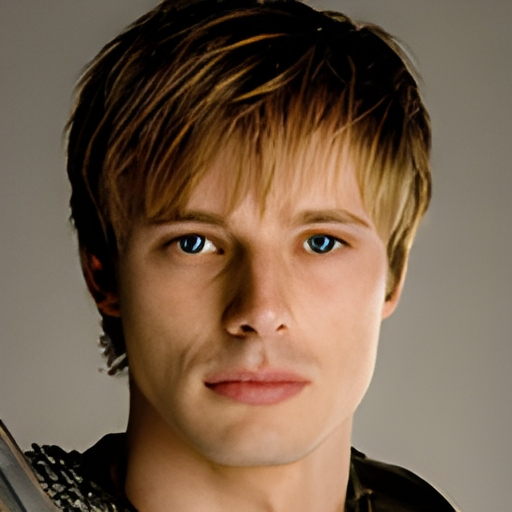} \\

\textbf{SDB} 
& \includegraphics[width=\linewidth]{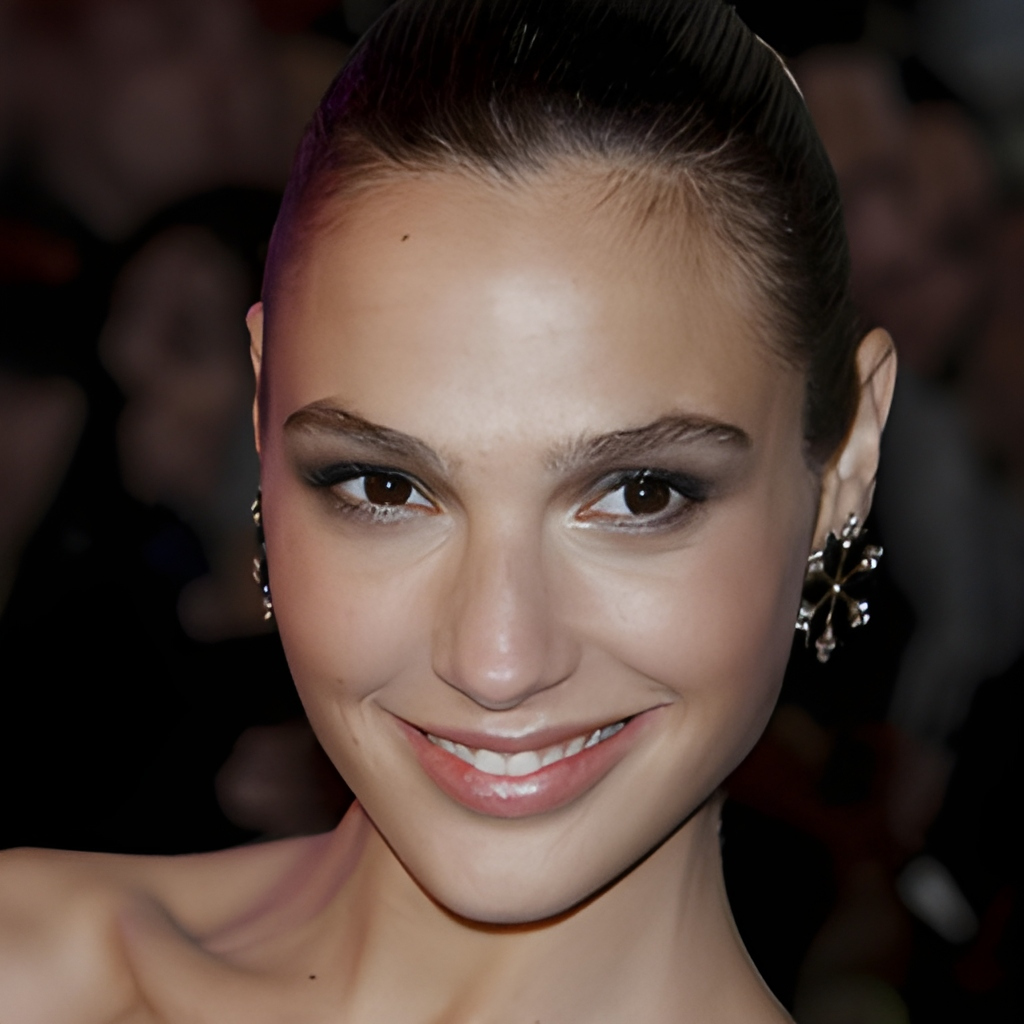} 
& \includegraphics[width=\linewidth]{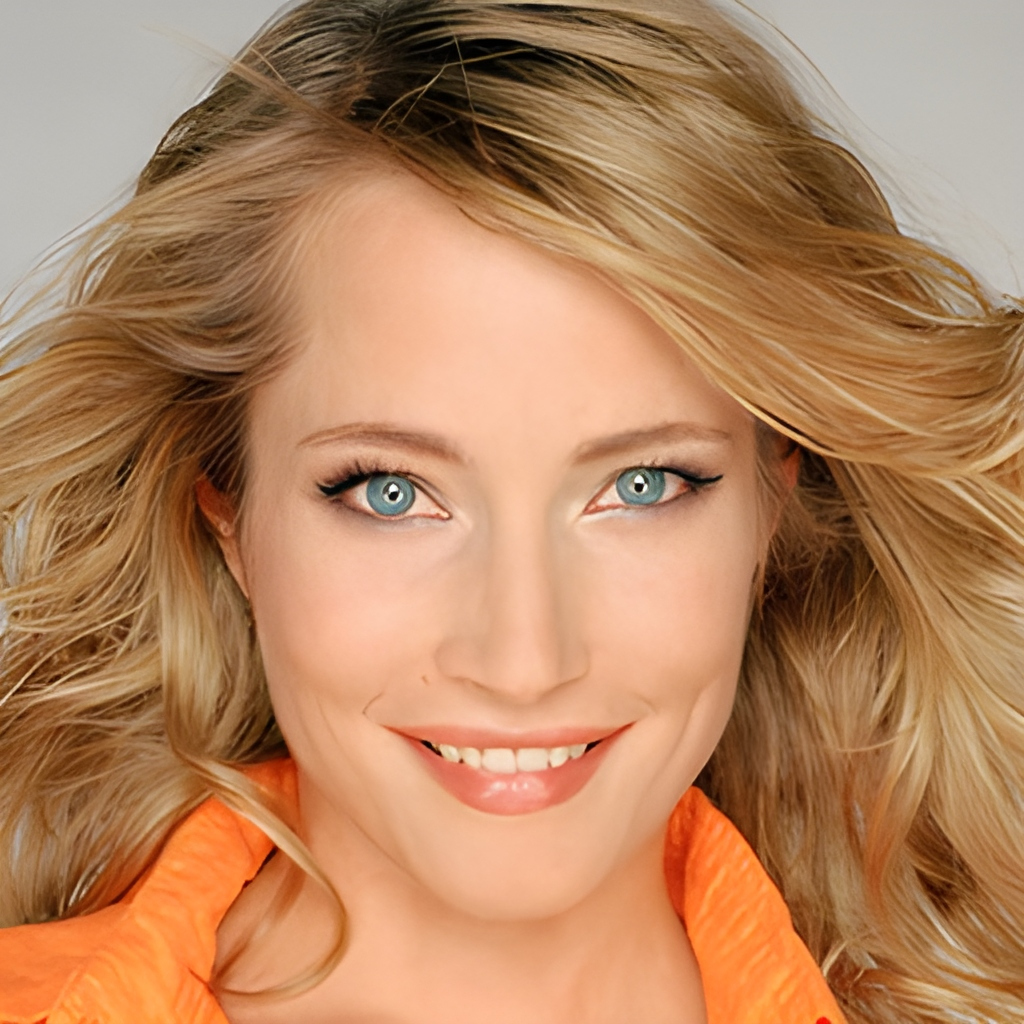} 
& \includegraphics[width=\linewidth]{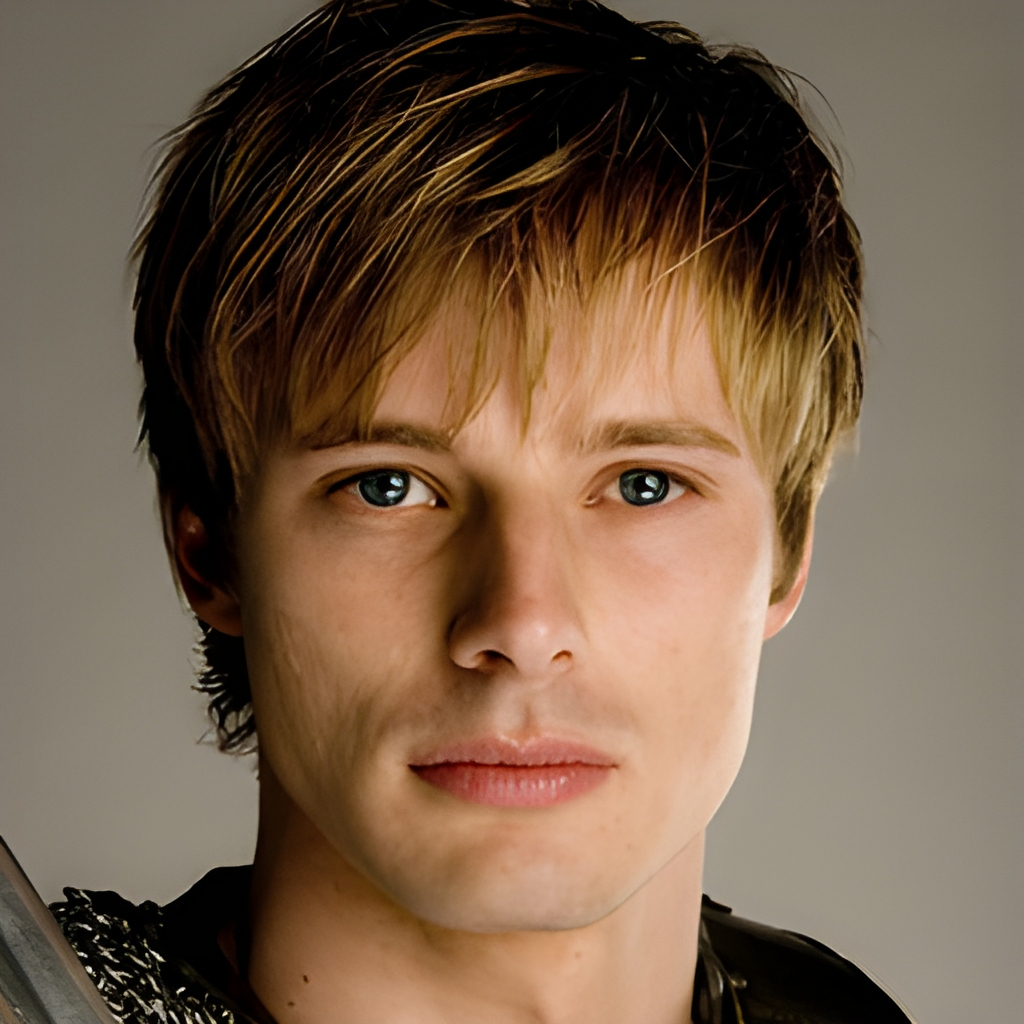} \\
\end{tabular}

\caption{Qualitative results on samples from the CelebA-HQ dataset. $\rho=1$.}
\label{fig:grid3x5_placeholders}
\end{figure}\newpage
\begin{figure}[h!]
\centering
\setlength{\tabcolsep}{10pt}
\renewcommand{\arraystretch}{1.15}

\begin{tabular}{>{\raggedleft\arraybackslash}m{1.8cm} *{3}{m{0.22\textwidth}}}

\textbf{Original} 
& \includegraphics[width=\linewidth]{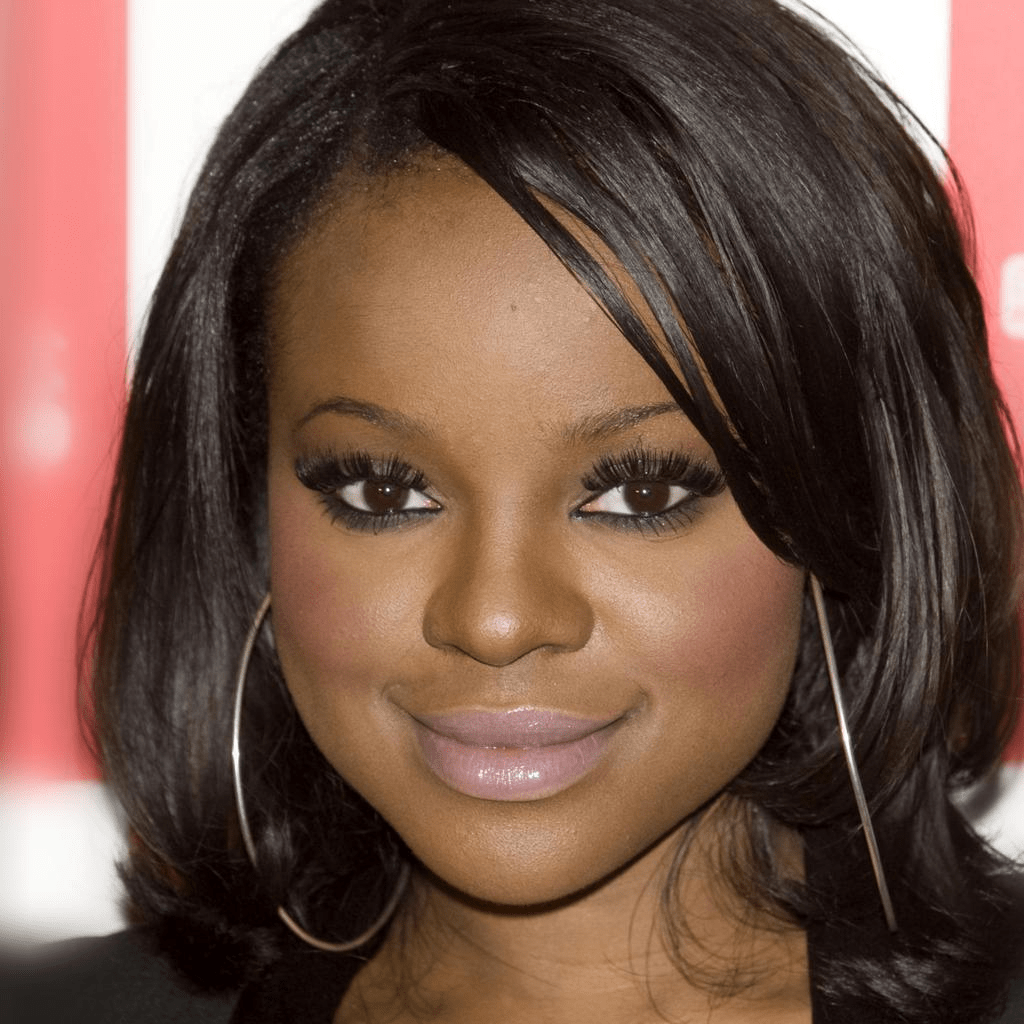} 
& \includegraphics[width=\linewidth]{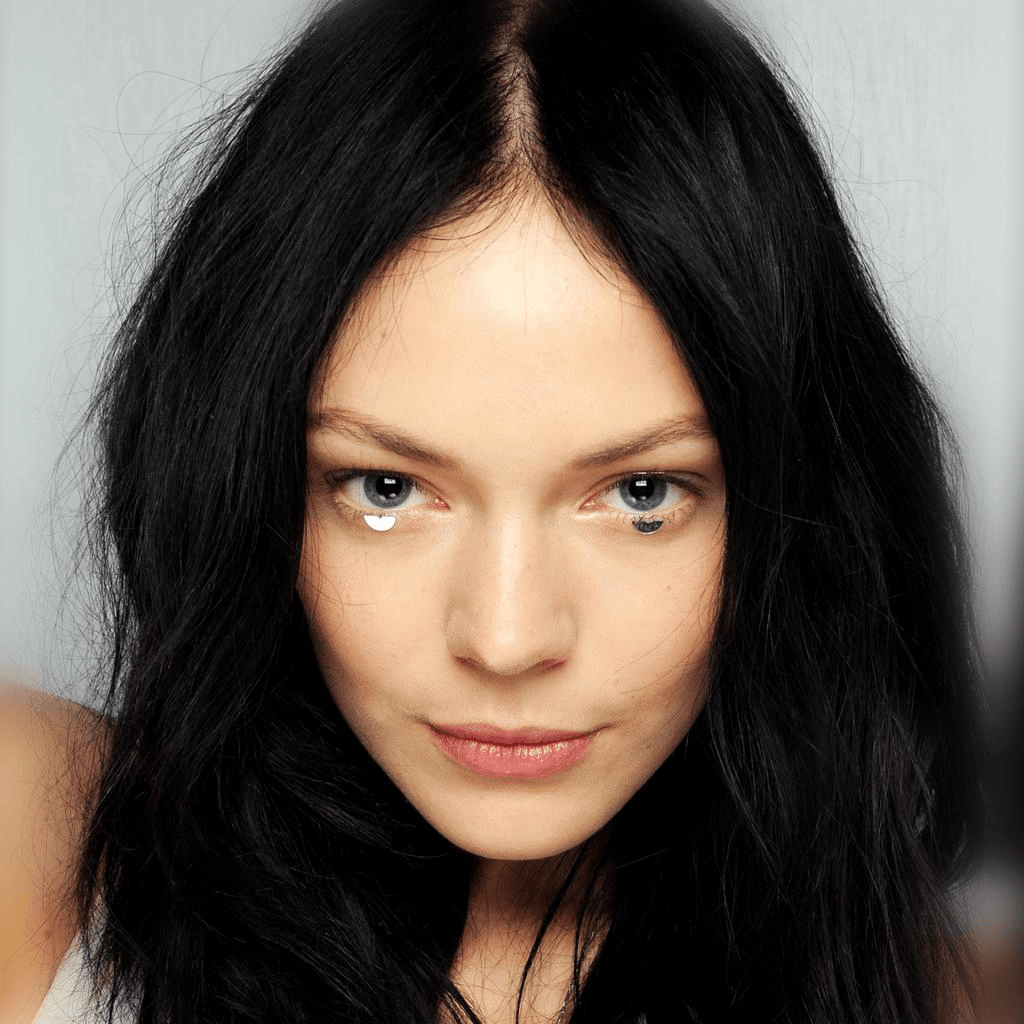} 
& \includegraphics[width=\linewidth]{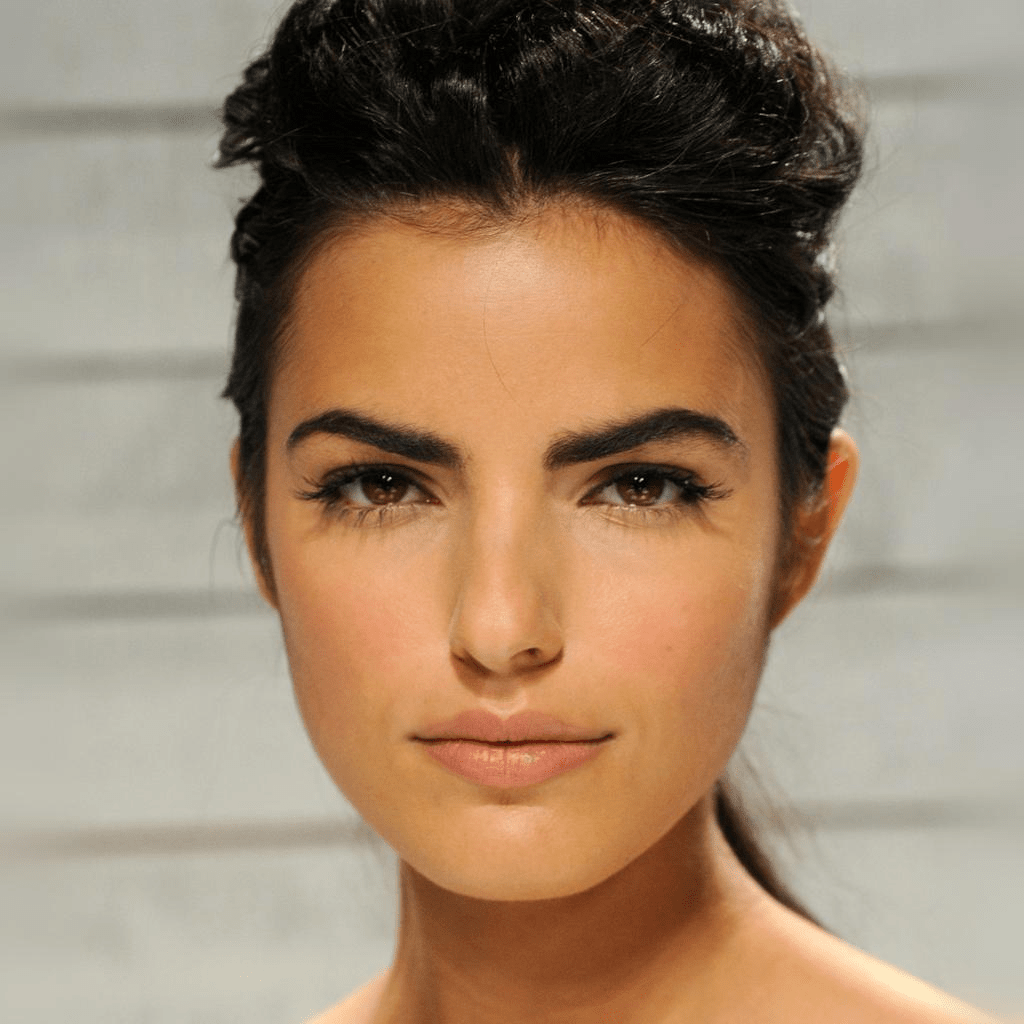} \\

\textbf{Down sampled} 
& \includegraphics[width=\linewidth]{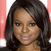} 
& \includegraphics[width=\linewidth]{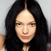} 
& \includegraphics[width=\linewidth]{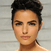} \\

\textbf{DiWa} 
& \includegraphics[width=\linewidth]{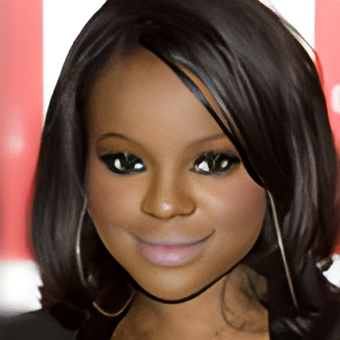} 
& \includegraphics[width=\linewidth]{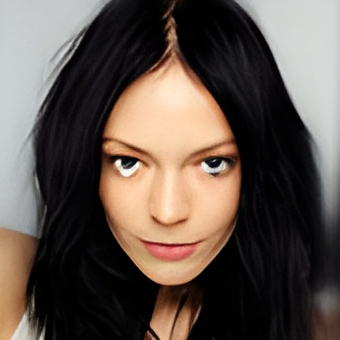} 
& \includegraphics[width=\linewidth]{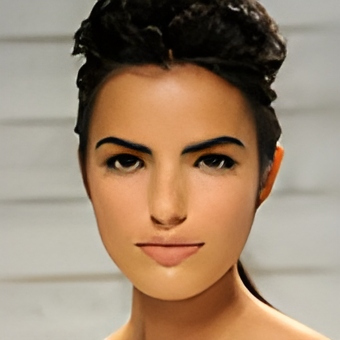} \\

\textbf{LDDBM} 

& \includegraphics[width=\linewidth]{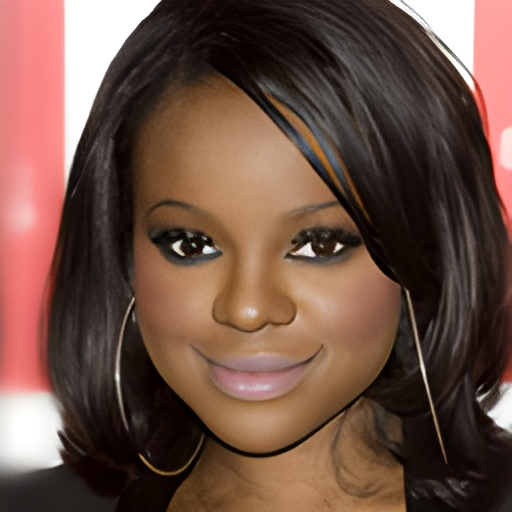} 
& \includegraphics[width=\linewidth]{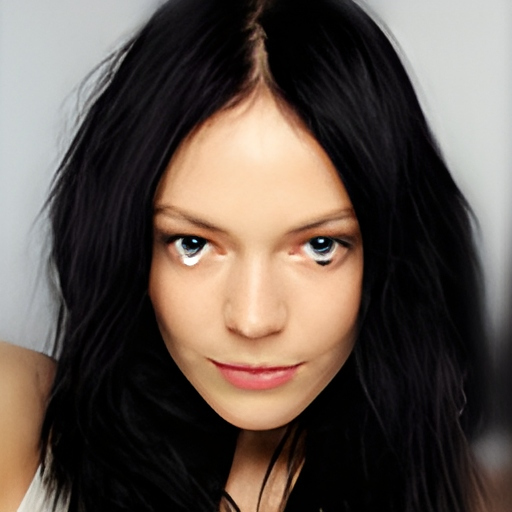} 
& \includegraphics[width=\linewidth]{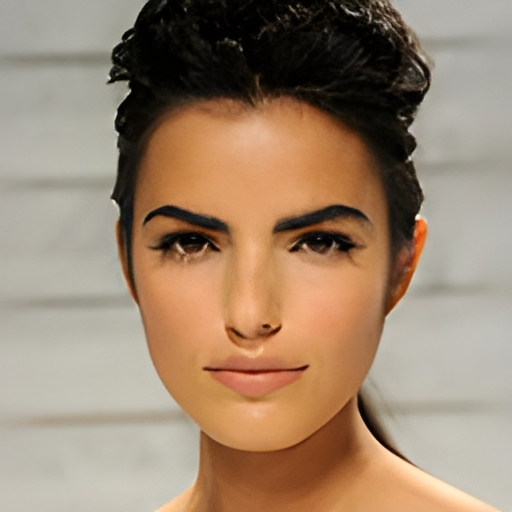} \\

\textbf{SDB} 
& \includegraphics[width=\linewidth]{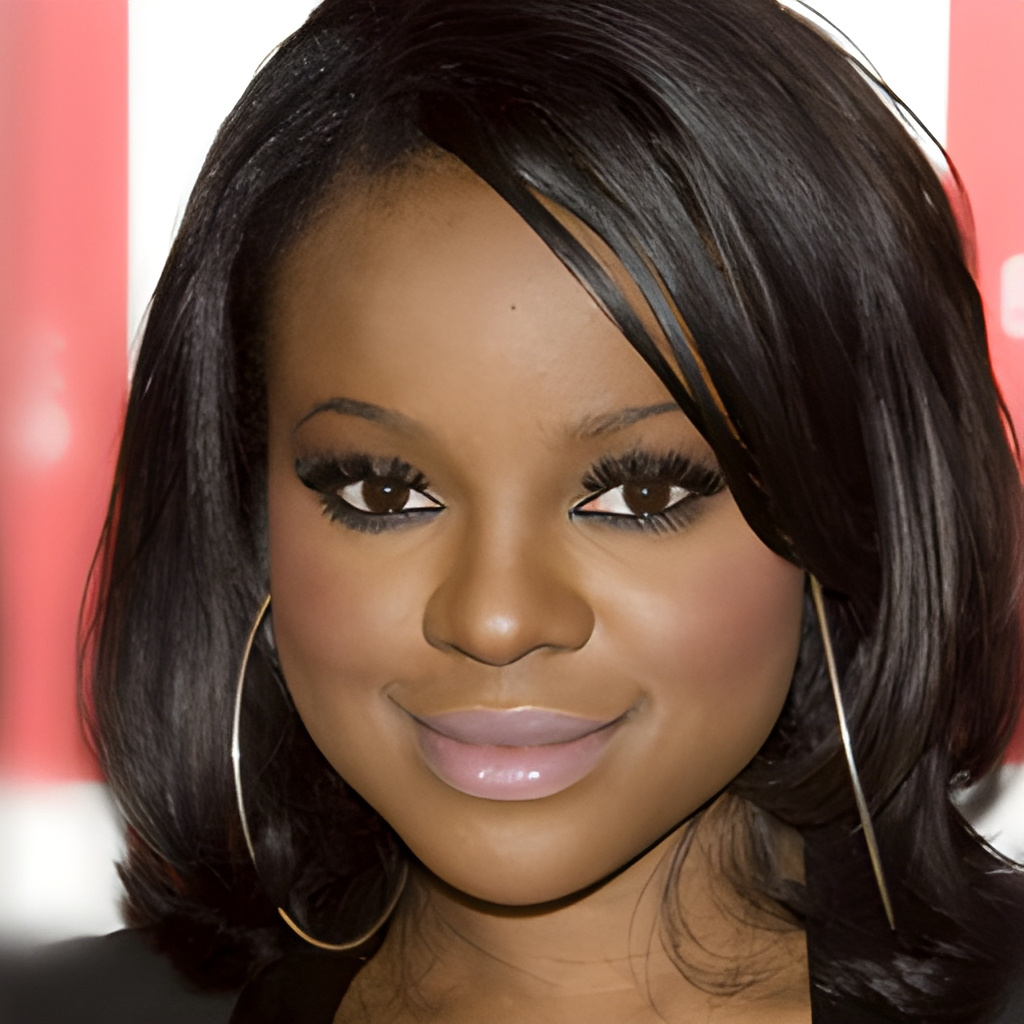} 
& \includegraphics[width=\linewidth]{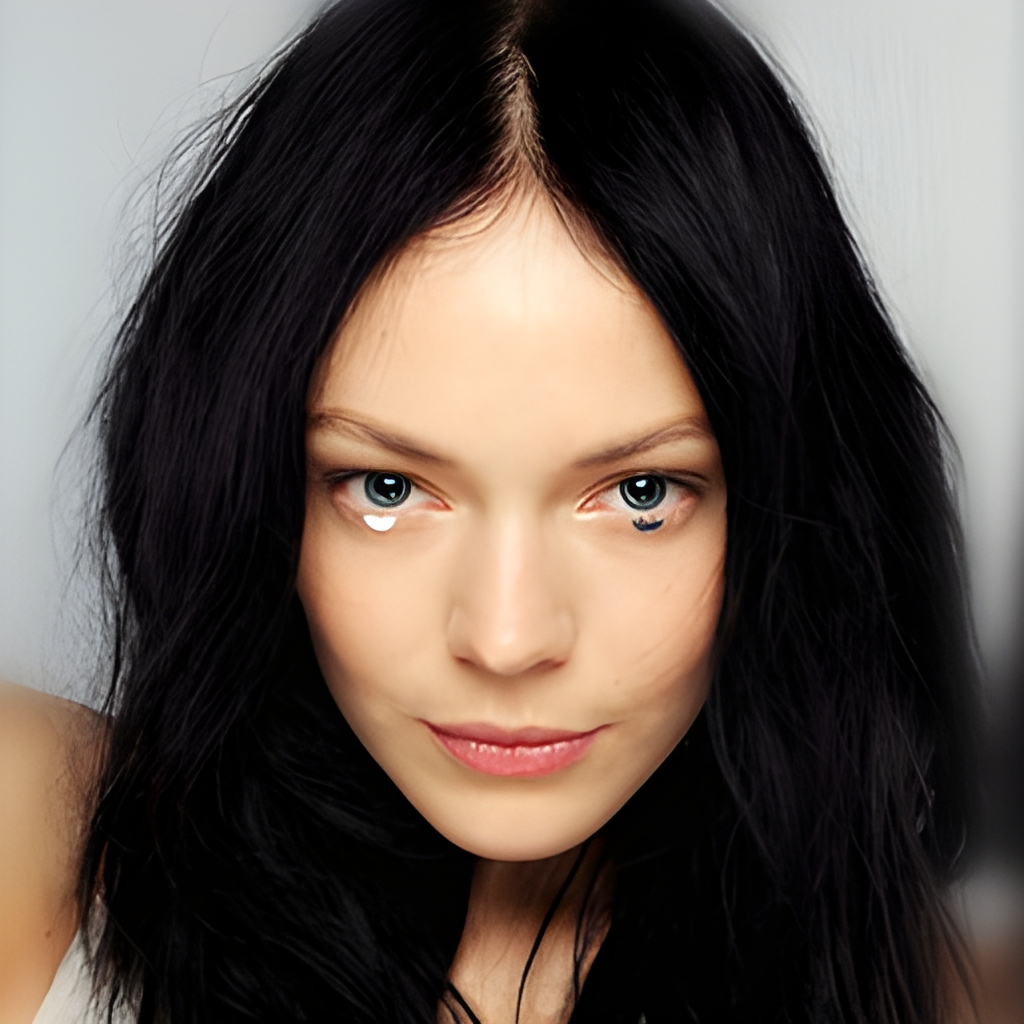} 
& \includegraphics[width=\linewidth]{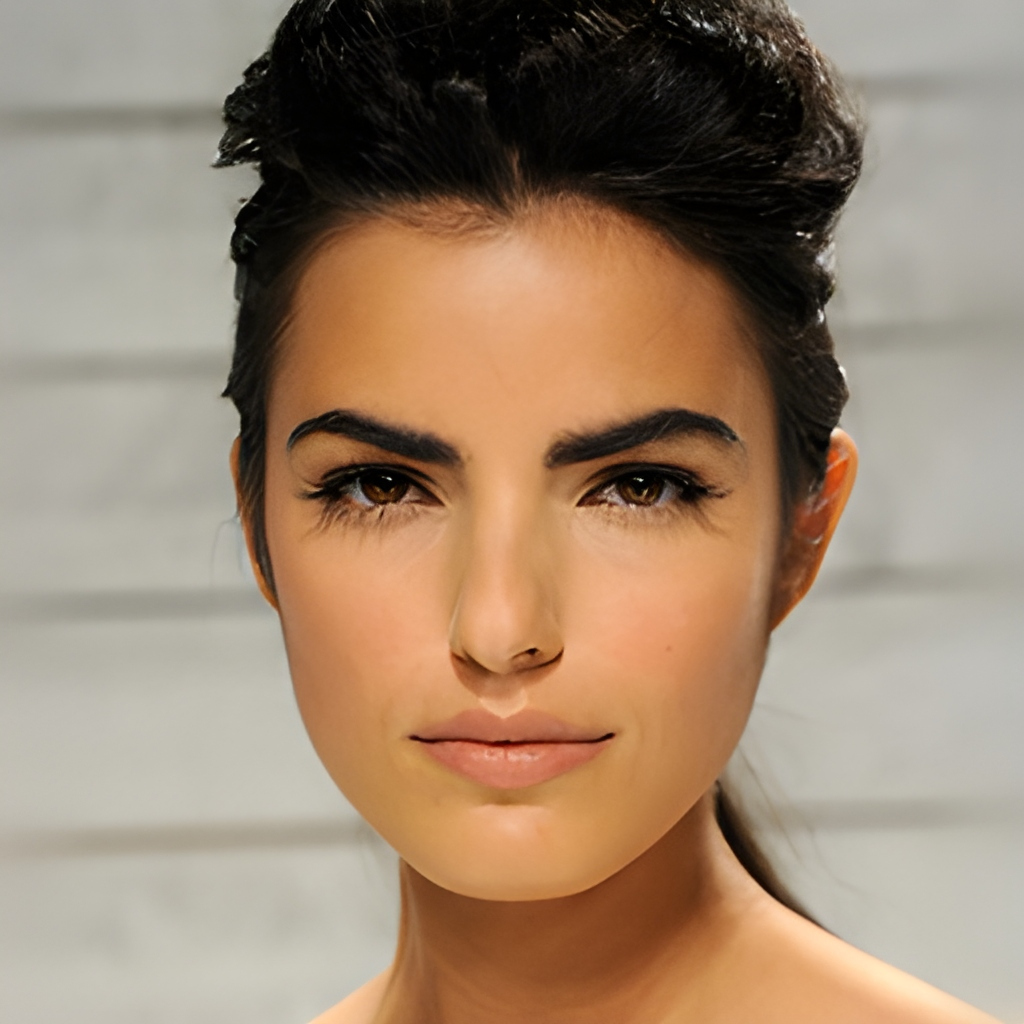} \\
\end{tabular}

\caption{Qualitative results on samples from the CelebA-HQ dataset. $\rho=1$.}
\label{fig:grid3x5_placeholders}
\end{figure}\newpage
\begin{figure}[h!]
\centering
\setlength{\tabcolsep}{10pt}
\renewcommand{\arraystretch}{1.15}

\begin{tabular}{>{\raggedleft\arraybackslash}m{1.8cm} *{3}{m{0.22\textwidth}}}

\textbf{Original} 
& \includegraphics[width=\linewidth]{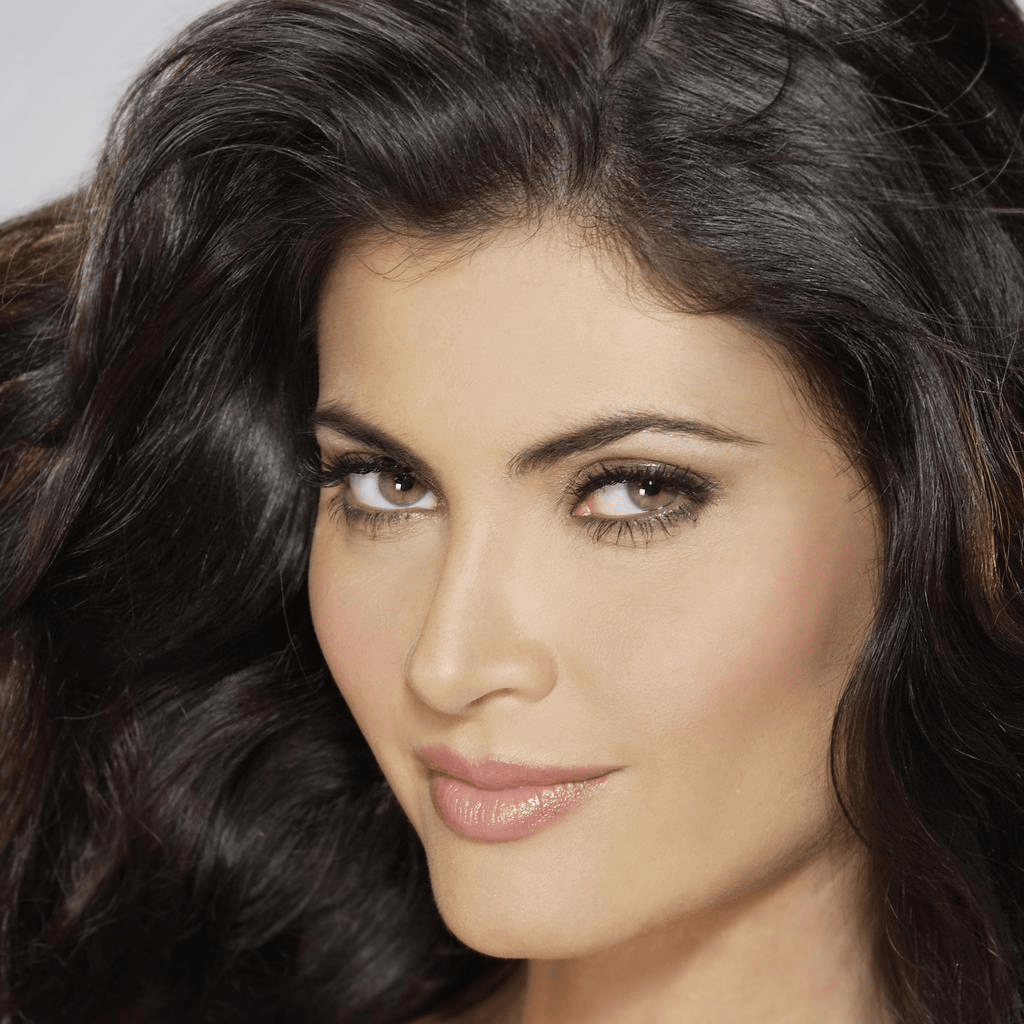} 
& \includegraphics[width=\linewidth]{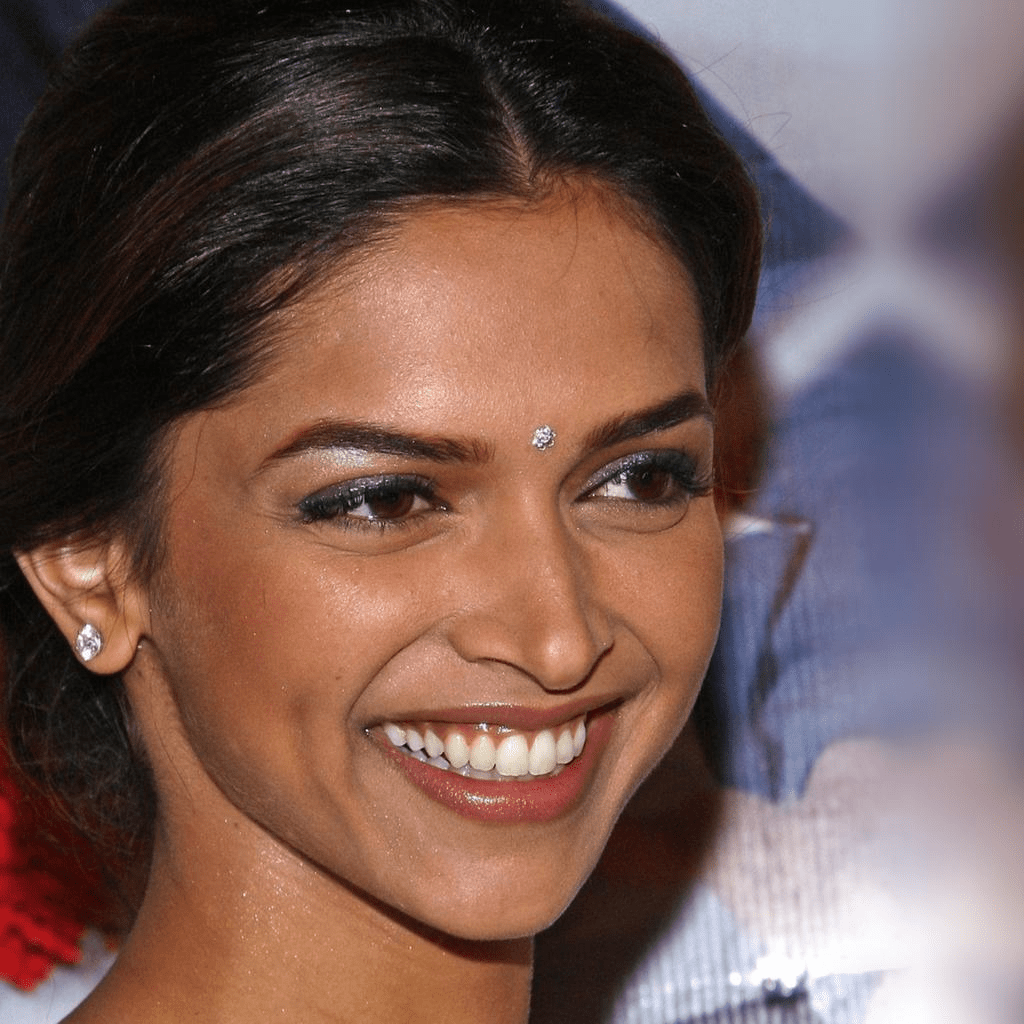} 
& \includegraphics[width=\linewidth]{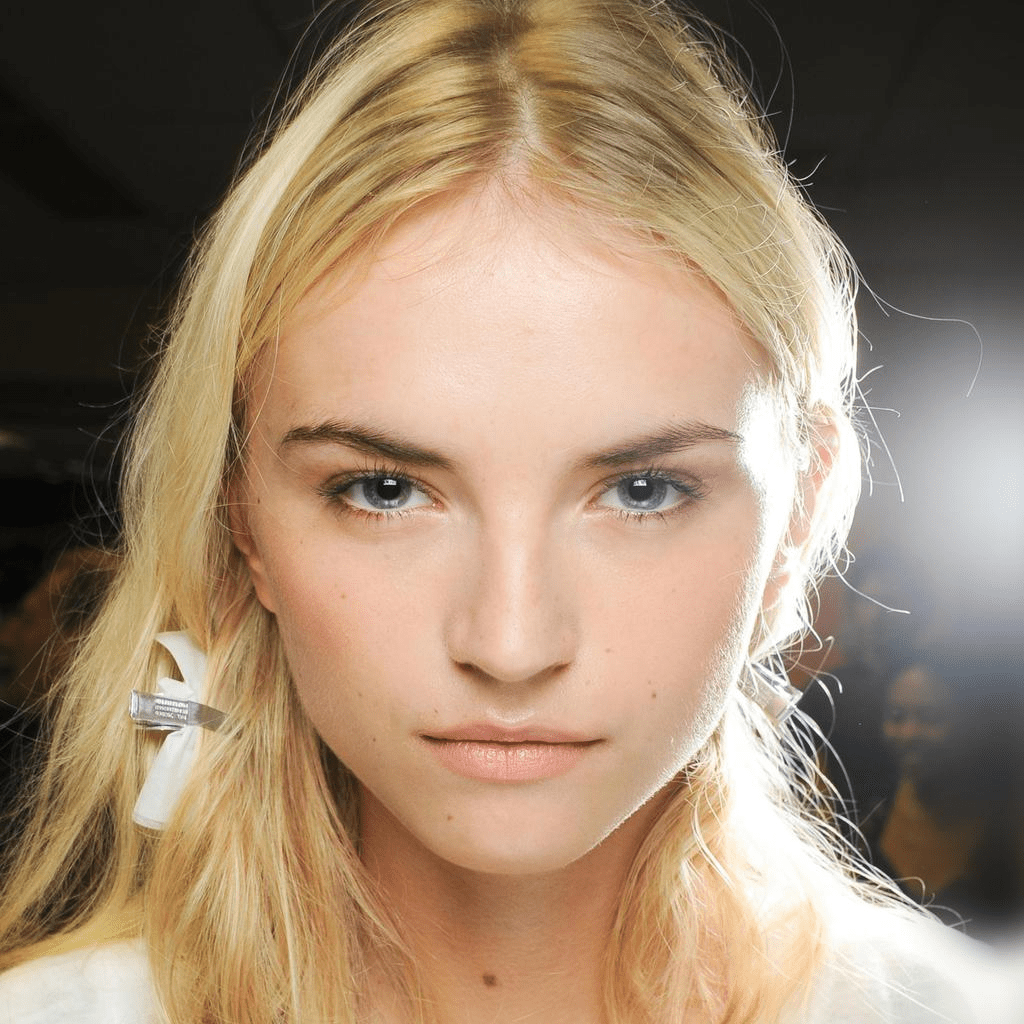} \\

\textbf{Down sampled} 
& \includegraphics[width=\linewidth]{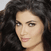} 
& \includegraphics[width=\linewidth]{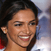} 
& \includegraphics[width=\linewidth]{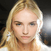} \\

\textbf{DiWa} 
& \includegraphics[width=\linewidth]{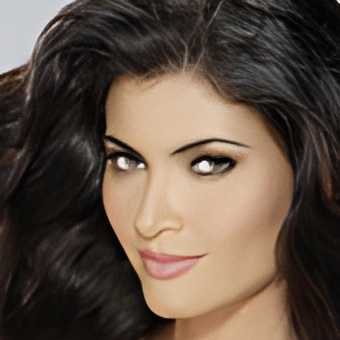} 
& \includegraphics[width=\linewidth]{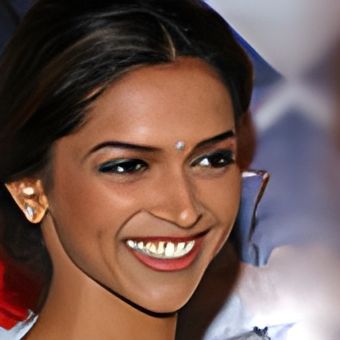} 
& \includegraphics[width=\linewidth]{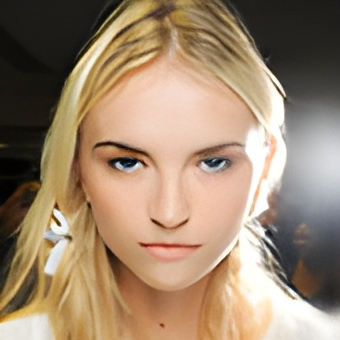} \\

\textbf{LDDBM} 

& \includegraphics[width=\linewidth]{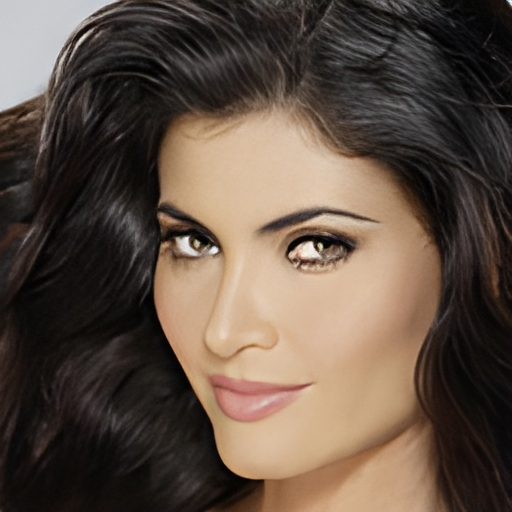} 
& \includegraphics[width=\linewidth]{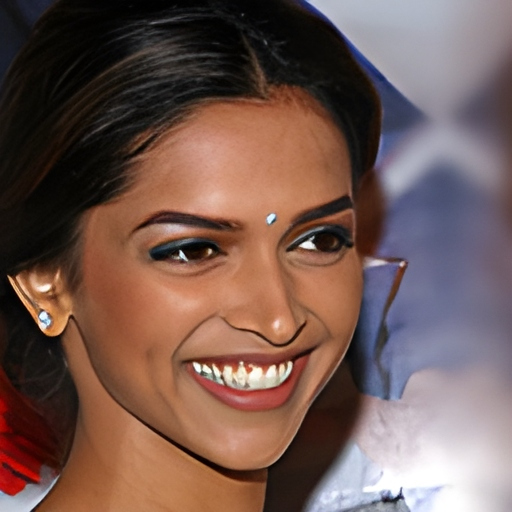} 
& \includegraphics[width=\linewidth]{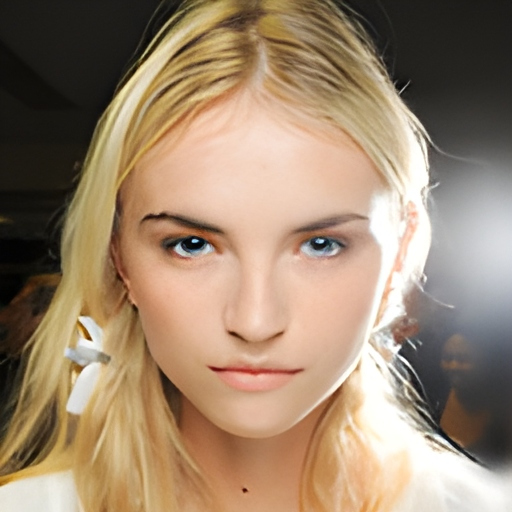} \\

\textbf{SDB} 
& \includegraphics[width=\linewidth]{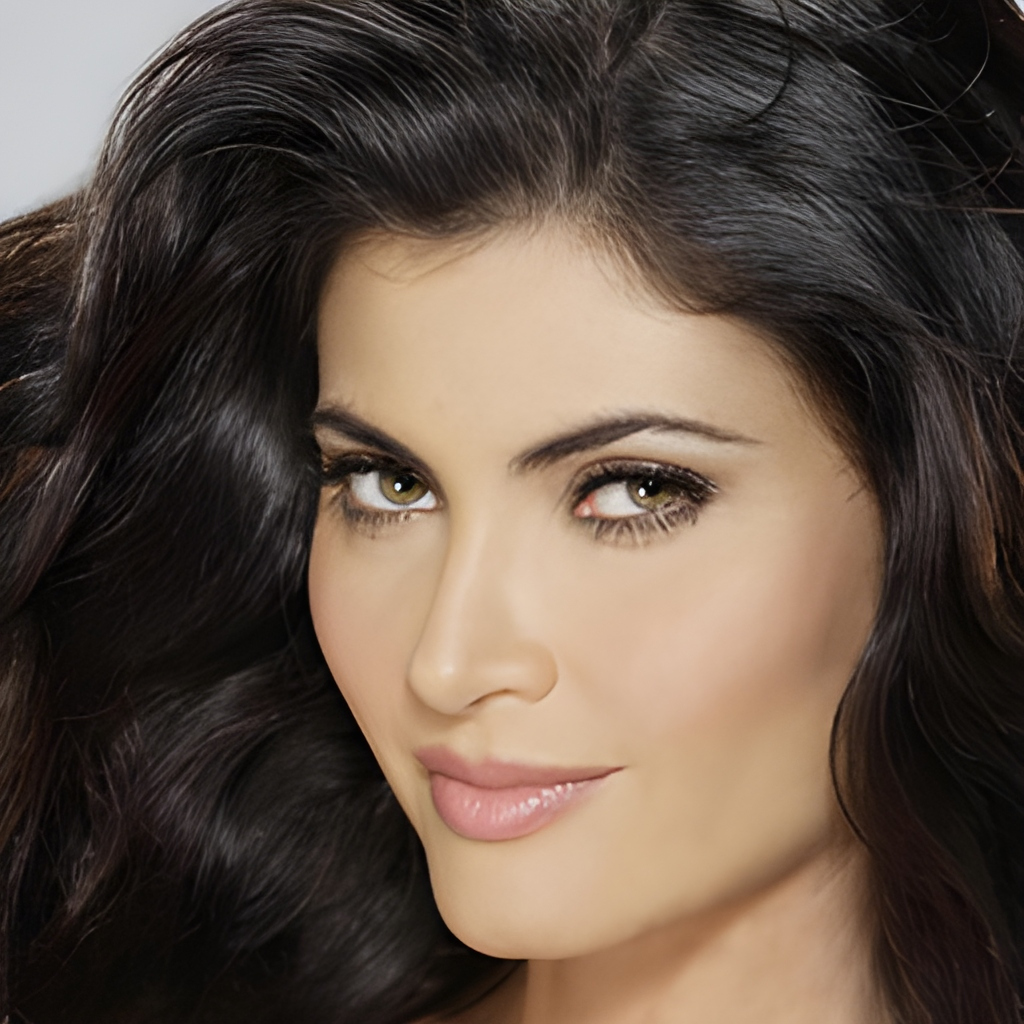} 
& \includegraphics[width=\linewidth]{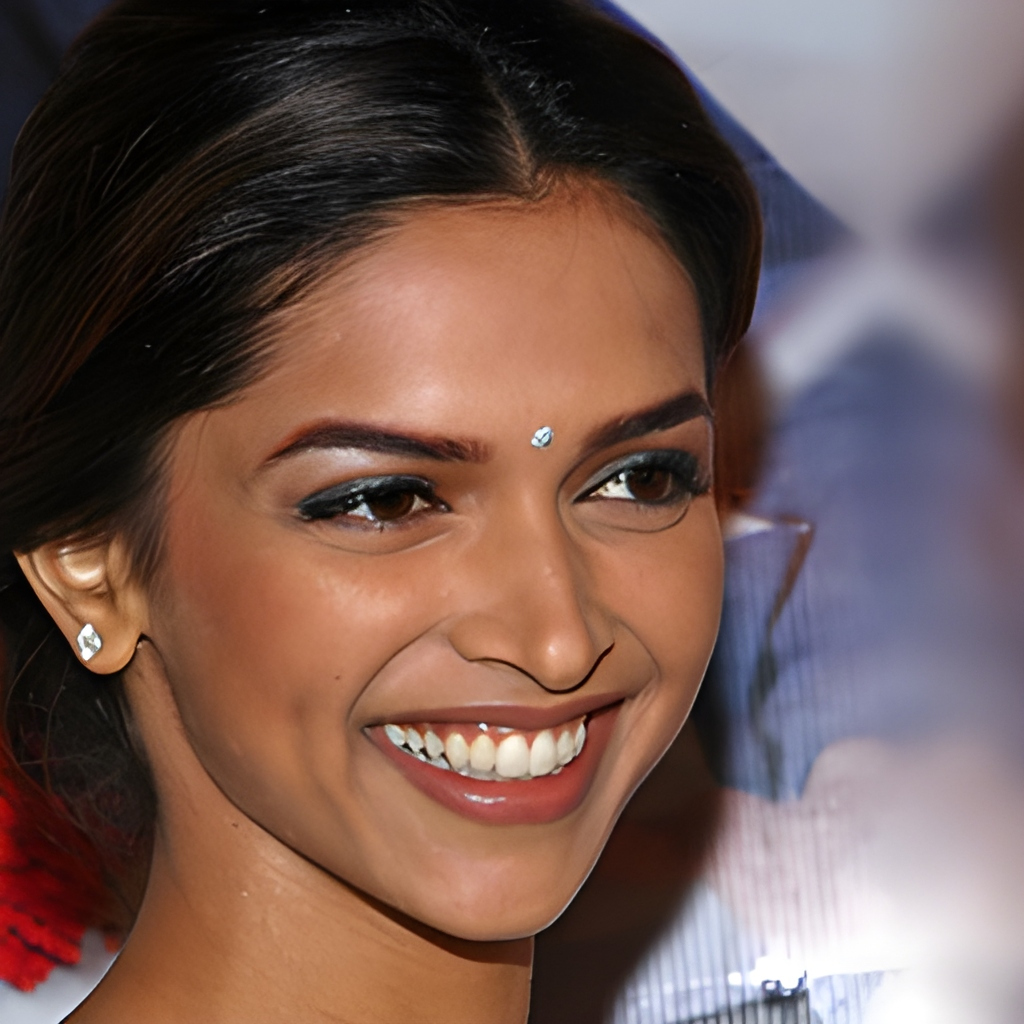} 
& \includegraphics[width=\linewidth]{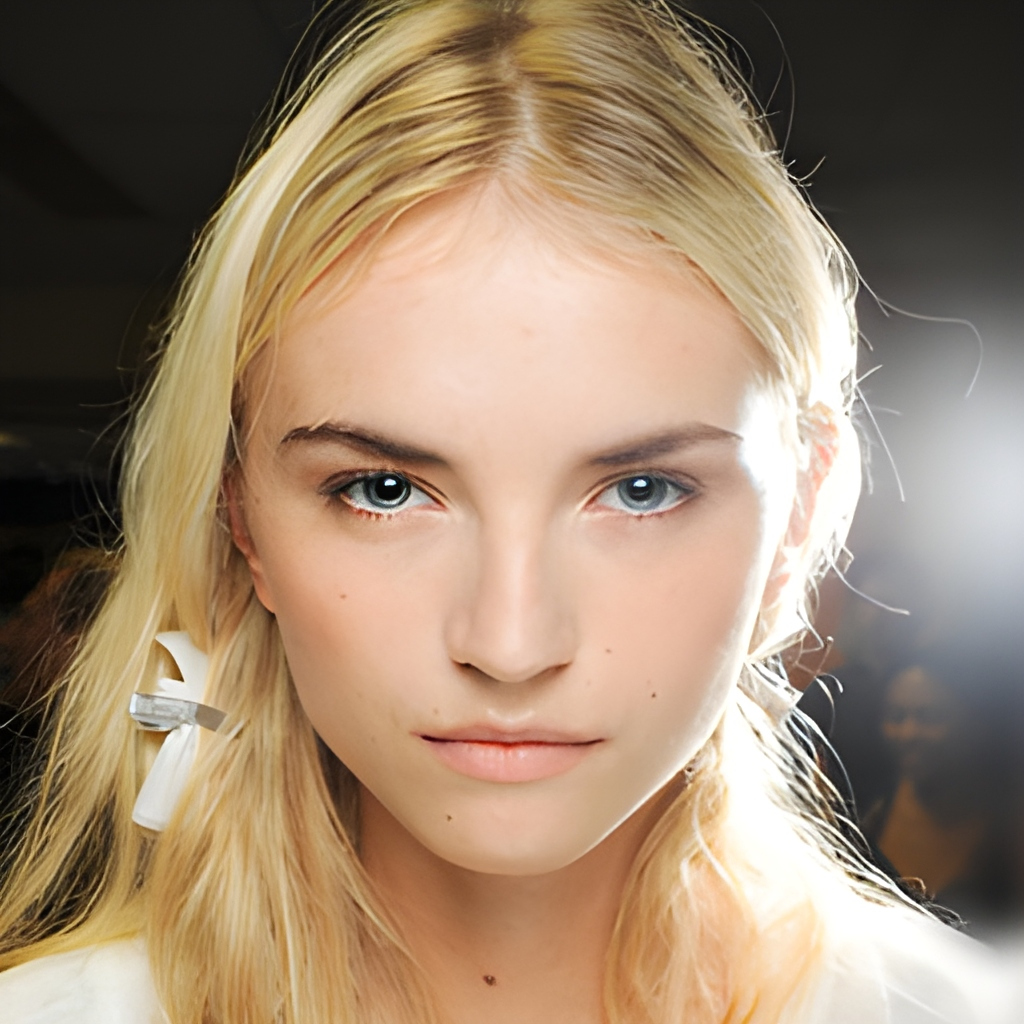} \\
\end{tabular}

\caption{Qualitative results on samples from the CelebA-HQ dataset. $\rho=1$.}
\label{fig:grid3x5_placeholders}
\end{figure}\newpage

\begin{figure}[h!]
\centering
\setlength{\tabcolsep}{10pt}
\renewcommand{\arraystretch}{1.15}

\begin{tabular}{>{\raggedleft\arraybackslash}m{1.8cm} *{3}{m{0.22\textwidth}}}

\textbf{Original} 
& \includegraphics[width=\linewidth]{results/sr/samples/original/0.png} 
& \includegraphics[width=\linewidth]{results/sr/samples/original/1.png} 
& \includegraphics[width=\linewidth]{results/sr/samples/original/2.png} \\

\textbf{Down sampled} 
& \includegraphics[width=\linewidth]{results/sr/samples/downsampled/0.png} 
& \includegraphics[width=\linewidth]{results/sr/samples/downsampled/1.png} 
& \includegraphics[width=\linewidth]{results/sr/samples/downsampled/2.png} \\

\textbf{DiWa} 
& \includegraphics[width=\linewidth]{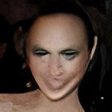} 
& \includegraphics[width=\linewidth]{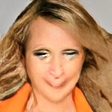} 
& \includegraphics[width=\linewidth]{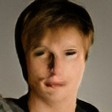} \\

\textbf{LDDBM} 

& \includegraphics[width=\linewidth]{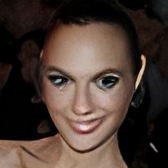} 
& \includegraphics[width=\linewidth]{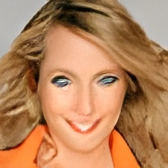} 
& \includegraphics[width=\linewidth]{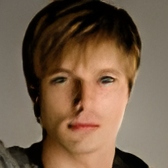} \\

\textbf{SDB} 
& \includegraphics[width=\linewidth]{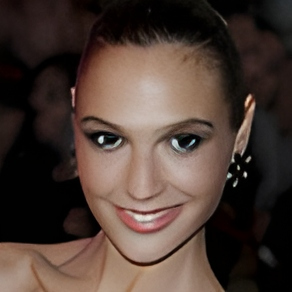} 
& \includegraphics[width=\linewidth]{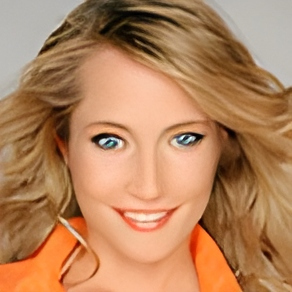} 
& \includegraphics[width=\linewidth]{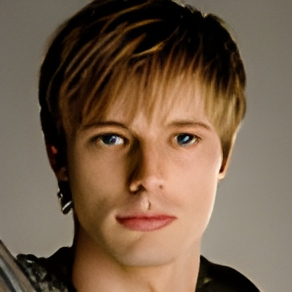} \\
\end{tabular}

\caption{Qualitative results on samples from the CelebA-HQ dataset. $\rho=0.5$.}
\label{fig:grid3x5_placeholders}
\end{figure}\newpage
\begin{figure}[h!]
\centering
\setlength{\tabcolsep}{10pt}
\renewcommand{\arraystretch}{1.15}

\begin{tabular}{>{\raggedleft\arraybackslash}m{1.8cm} *{3}{m{0.22\textwidth}}}

\textbf{Original} 
& \includegraphics[width=\linewidth]{results/sr/samples/original/3.png} 
& \includegraphics[width=\linewidth]{results/sr/samples/original/4.png} 
& \includegraphics[width=\linewidth]{results/sr/samples/original/5.png} \\

\textbf{Down sampled} 
& \includegraphics[width=\linewidth]{results/sr/samples/downsampled/3.png} 
& \includegraphics[width=\linewidth]{results/sr/samples/downsampled/4.png} 
& \includegraphics[width=\linewidth]{results/sr/samples/downsampled/5.png} \\

\textbf{DiWa} 
& \includegraphics[width=\linewidth]{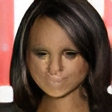} 
& \includegraphics[width=\linewidth]{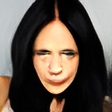} 
& \includegraphics[width=\linewidth]{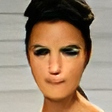} \\

\textbf{LDDBM} 

& \includegraphics[width=\linewidth]{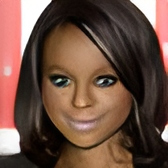} 
& \includegraphics[width=\linewidth]{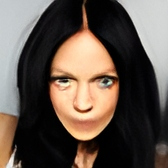} 
& \includegraphics[width=\linewidth]{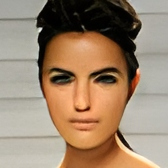} \\

\textbf{SDB} 
& \includegraphics[width=\linewidth]{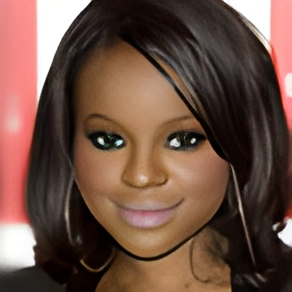} 
& \includegraphics[width=\linewidth]{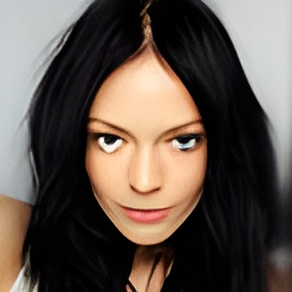} 
& \includegraphics[width=\linewidth]{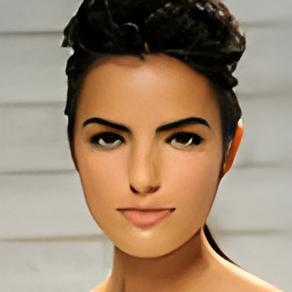} \\
\end{tabular}

\caption{Qualitative results on samples from the CelebA-HQ dataset. $\rho=0.5$.}
\label{fig:grid3x5_placeholders}
\end{figure}\newpage
\begin{figure}[h!]
\centering
\setlength{\tabcolsep}{10pt}
\renewcommand{\arraystretch}{1.15}

\begin{tabular}{>{\raggedleft\arraybackslash}m{1.8cm} *{3}{m{0.22\textwidth}}}

\textbf{Original} 
& \includegraphics[width=\linewidth]{results/sr/samples/original/6.png} 
& \includegraphics[width=\linewidth]{results/sr/samples/original/7.png} 
& \includegraphics[width=\linewidth]{results/sr/samples/original/8.png} \\

\textbf{Down sampled} 
& \includegraphics[width=\linewidth]{results/sr/samples/downsampled/6.png} 
& \includegraphics[width=\linewidth]{results/sr/samples/downsampled/7.png} 
& \includegraphics[width=\linewidth]{results/sr/samples/downsampled/8.png} \\

\textbf{DiWa} 
& \includegraphics[width=\linewidth]{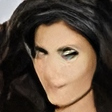} 
& \includegraphics[width=\linewidth]{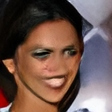} 
& \includegraphics[width=\linewidth]{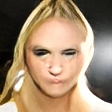} \\

\textbf{LDDBM} 

& \includegraphics[width=\linewidth]{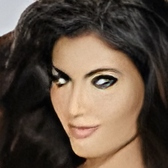} 
& \includegraphics[width=\linewidth]{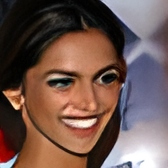} 
& \includegraphics[width=\linewidth]{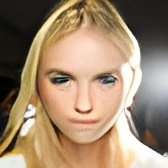} \\

\textbf{SDB} 
& \includegraphics[width=\linewidth]{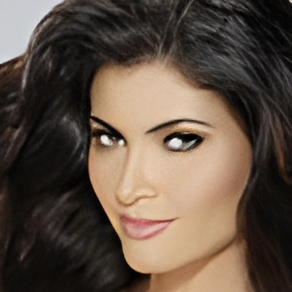} 
& \includegraphics[width=\linewidth]{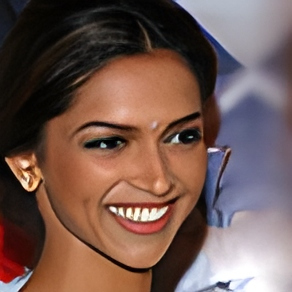} 
& \includegraphics[width=\linewidth]{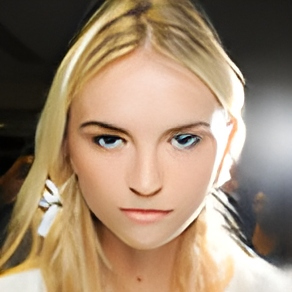} \\
\end{tabular}

\caption{Qualitative results on samples from the CelebA-HQ dataset. $\rho=0.5$.}
\label{fig:grid3x5_placeholders}
\end{figure}\newpage

\begin{figure}[h!]
\centering
\setlength{\tabcolsep}{10pt}
\renewcommand{\arraystretch}{1.15}

\begin{tabular}{>{\raggedleft\arraybackslash}m{1.8cm} *{3}{m{0.22\textwidth}}}

\textbf{Original} 
& \includegraphics[width=\linewidth]{results/sr/samples/original/0.png} 
& \includegraphics[width=\linewidth]{results/sr/samples/original/1.png} 
& \includegraphics[width=\linewidth]{results/sr/samples/original/2.png} \\

\textbf{Down sampled} 
& \includegraphics[width=\linewidth]{results/sr/samples/downsampled/0.png} 
& \includegraphics[width=\linewidth]{results/sr/samples/downsampled/1.png} 
& \includegraphics[width=\linewidth]{results/sr/samples/downsampled/2.png} \\

\textbf{SDB} 
& \includegraphics[width=\linewidth]{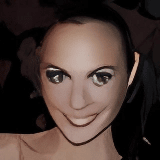} 
& \includegraphics[width=\linewidth]{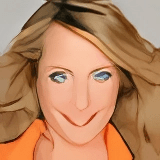} 
& \includegraphics[width=\linewidth]{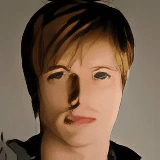} \\
\end{tabular}

\caption{Qualitative results on samples from the CelebA-HQ dataset. $\rho=0$.}
\label{fig:grid3x5_placeholders}
\end{figure}\newpage
\begin{figure}[h!]
\centering
\setlength{\tabcolsep}{10pt}
\renewcommand{\arraystretch}{1.15}

\begin{tabular}{>{\raggedleft\arraybackslash}m{1.8cm} *{3}{m{0.22\textwidth}}}

\textbf{Original} 
& \includegraphics[width=\linewidth]{results/sr/samples/original/3.png} 
& \includegraphics[width=\linewidth]{results/sr/samples/original/4.png} 
& \includegraphics[width=\linewidth]{results/sr/samples/original/5.png} \\

\textbf{Down sampled} 
& \includegraphics[width=\linewidth]{results/sr/samples/downsampled/3.png} 
& \includegraphics[width=\linewidth]{results/sr/samples/downsampled/4.png} 
& \includegraphics[width=\linewidth]{results/sr/samples/downsampled/5.png} \\

\textbf{SDB} 
& \includegraphics[width=\linewidth]{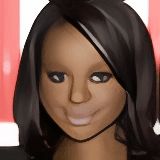} 
& \includegraphics[width=\linewidth]{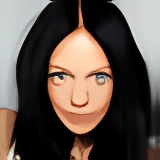} 
& \includegraphics[width=\linewidth]{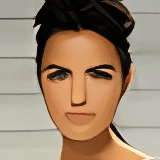} \\
\end{tabular}

\caption{Qualitative results on samples from the CelebA-HQ dataset. $\rho=0$.}
\label{fig:grid3x5_placeholders}
\end{figure}\newpage
\begin{figure}[h!]
\centering
\setlength{\tabcolsep}{10pt}
\renewcommand{\arraystretch}{1.15}

\begin{tabular}{>{\raggedleft\arraybackslash}m{1.8cm} *{3}{m{0.22\textwidth}}}

\textbf{Original} 
& \includegraphics[width=\linewidth]{results/sr/samples/original/6.png} 
& \includegraphics[width=\linewidth]{results/sr/samples/original/7.png} 
& \includegraphics[width=\linewidth]{results/sr/samples/original/8.png} \\

\textbf{Down sampled} 
& \includegraphics[width=\linewidth]{results/sr/samples/downsampled/6.png} 
& \includegraphics[width=\linewidth]{results/sr/samples/downsampled/7.png} 
& \includegraphics[width=\linewidth]{results/sr/samples/downsampled/8.png} \\

\textbf{SDB} 
& \includegraphics[width=\linewidth]{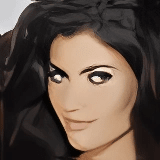} 
& \includegraphics[width=\linewidth]{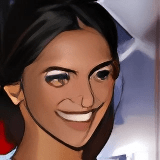} 
& \includegraphics[width=\linewidth]{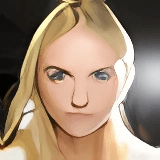} \\
\end{tabular}

\caption{Qualitative results on samples from the CelebA-HQ dataset. $\rho=0$.}
\label{fig:grid3x5_placeholders}
\end{figure}\newpage

\newpage
\section{Multi-view Images to 3D Voxel Grids Experimental Details}
\label{app:exp:mv2vox}

\subsection{Architectures}
\label{app:mv2vox_arch}

\paragraph{\textbf{Score model.}}
Similar to the Super-Resolution experiment, we use DTDiT architecture as in LDDBM \cite{berman2025towards} for the conditional bridge \(s_\theta(z_t,t\mid z_T)\), and unless stated otherwise, we match the LDDBM training budget and optimization settings for this benchmark.

\paragraph{\textbf{Autoencoders.}}
For a fair comparison, we adopt the architectures used in LDDBM \cite{berman2025towards}, shortly described below.

\begin{itemize}
\item \textbf{3D convolutional encoder.}
Encodes voxel inputs \(x \in \mathbb{R}^{B \times C \times D \times H \times W}\) using three 3D convolutional layers (kernel \(4\), stride \(2\), padding \(1\)) with channel progression \(C \rightarrow 64 \rightarrow 128 \rightarrow 256\). Each layer is followed by SiLU and optional dropout. The output is reshaped into a latent tensor used by the bridge model.

\item \textbf{3D convolutional decoder.}
Mirrors the 3D encoder using transposed 3D convolutions (kernel \(4\), stride \(2\), padding \(1\)) to upsample the latent representation back to the original voxel resolution, with SiLU activations and optional dropout.

\item \textbf{Multi-view 2D convolutional encoder.}
Encodes multi-view images \(y \in \mathbb{R}^{B \times V \times C \times H \times W}\) by applying a shared 2D convolutional stack to each view (flattening \(B \times V\)). The stack has six layers (kernel \(4\), stride \(2\), appropriate padding) with SiLU and optional dropout, increasing channels from \(32\) up to \(256\) while reducing spatial resolution. Per-view features are concatenated across views and projected with a fully connected layer to obtain a unified latent representation.

\item \textbf{Identity encoder/decoder (3D-EDM baseline).}
For the 3D-EDM baseline, diffusion operates directly in voxel space, so the encoder and decoder are identity maps (no latent-space adaptation).
\end{itemize}

\subsection{Evaluation Protocol}
\label{sec:mv2vox_eval_protocol}

We use the ShapeNet \cite{wu20153d} dataset which comprises tens of thousands of 3D models across 55 object categories. We use the standard subset of 13 categories (approximately 43,783 models). Each 3D model is voxelized using Binvox at resolution \(32 \times 32 \times 32\), and we render \(V=4\) images per model from random viewpoints at resolution \(224 \times 224\). We split objects into 70\%/10\%/20\% train/validation/test, maintaining a balanced distribution across categories.

For comparison, we include EDM and EDM-3D \cite{karras2022elucidating} as diffusion-based baselines using 2D and 3D U-Net architectures \cite{ronneberger2015u}, respectively. Additionally, we report LRGT+ \cite{yang2023longrangegroupingtransformermultiview} as a representative task-specialized baseline to illustrate the trade-off between general-purpose translation and domain-specific inductive biases.

\subsection{Metrics}
\label{sec:mv2vox_metrics}

We evaluate the voxel generation using both generative and reconstruction-based metrics. We report 1-nearest neighbor accuracy (1-NNA) as a distributional similarity test between generated and real shapes, and additionally report voxel Intersection-over-Union (IoU) to assess instance-level reconstruction fidelity. Distances for 1-NNA are computed using Chamfer Distance (CD) between point clouds.

\begin{itemize}
\item \textbf{Chamfer Distance (CD).}
To compare 3D shapes for 1-NNA, we convert voxel grids into point clouds and compute Chamfer Distance. Given two point clouds \(P\) and \(Q\),
\[
\mathrm{CD}(P,Q) \;=\; \frac{1}{|P|}\sum_{p\in P}\min_{q\in Q}\|p-q\|_2^2 \;+\; \frac{1}{|Q|}\sum_{q\in Q}\min_{p\in P}\|q-p\|_2^2.
\]
Lower CD indicates higher geometric similarity. We use CD only as the distance function underlying 1-NNA and do not report it as a standalone metric.

\item \textbf{Distributional similarity (1-NNA).}
1-nearest neighbor accuracy (1-NNA) is a two-sample test that assesses whether generated samples are distinguishable from real samples \citep{yang2019pointflow, zhou20213d}. Let \(S_{\mathrm{real}}\) and \(S_{\mathrm{gen}}\) denote sets of real and generated samples, and define \(S = S_{\mathrm{real}} \cup S_{\mathrm{gen}}\) with binary labels indicating sample origin. For each \(x\in S\), we find its nearest neighbor \(\mathrm{NN}(x)\) in \(S\setminus\{x\}\) using Chamfer Distance, and perform leave-one-out 1-NN classification. The score is
\[
\mathrm{1\mbox{-}NNA} \;=\; \frac{1}{|S|}\sum_{x\in S}\mathds{1}\!\left[\mathrm{label}(x)=\mathrm{label}(\mathrm{NN}(x))\right].
\]
An ideal generative model yields samples indistinguishable from real data, giving \(\mathrm{1\mbox{-}NNA}\approx 0.5\). Values significantly above \(0.5\) indicate that the distributions are separable; thus lower is better, with scores closer to \(0.5\) reflecting stronger generative performance.

\item \textbf{Reconstruction fidelity (IoU).}
We compute Intersection-over-Union (IoU) between a predicted occupancy grid \(\hat{V}\in\{0,1\}^{R\times R\times R}\) and the ground-truth grid \(V\in\{0,1\}^{R\times R\times R}\),
\[
\mathrm{IoU} \;=\; \frac{|\hat{V}\cap V|}{|\hat{V}\cup V|}.
\]
Higher IoU indicates better overlap with the true shape and is most relevant when the goal is to reconstruct a specific instance from the conditioning views.
\end{itemize}

Since the data are voxelized, we first convert voxel grids to point clouds to compute Chamfer distances for 1-NNA, and then follow the standard evaluation interface of \cite{shi20213d} to obtain the final scores.

\subsubsection{Real-data ablation of structural constraints}
\label{app:mv2vox_ablation}
Analogously to the super-resolution ablation in Appendix~\ref{app:sr_ablation}, we provide a
component-wise ablation for multi-view-to-3D translation in Table~\ref{tab:shapenets_ablation}. Rows that do
not include the paired objective are independent of $\rho$ and are repeated
across supervision blocks to make the comparison with paired-only and
semi-paired objectives explicit. The results show that MM alone improves
distributional validity but gives weaker reconstruction fidelity, endpoint CC
improves both 1-NNA and IoU, and trajectory-level CC gives the strongest
unpaired structural baseline. Adding paired supervision on top of these
constraints gives the best overall performance.

\begin{table*}[h]
\centering
\caption{\textbf{Real-data ablation of SDB constraints on multi-view-to-3D translation.}
Rows without the paired objective are independent of $\rho$ and are repeated for
comparison with paired-only and semi-paired training. We report distributional
similarity using 1-NNA and reconstruction fidelity using IoU. Ideal 1-NNA is
approximately $0.5$, and higher IoU indicates better instance-level accuracy.}
\label{tab:shapenets_ablation}
\begin{tabular}{llcc}
\toprule
$\rho$ & Method
& 1-NNA $\downarrow$
& IoU $\uparrow$ \\
\midrule
\multirow{3}{*}{0}
& Marginal matching only
& $0.706{\pm}0.010$ & $0.476{\pm}0.010$ \\
& + Endpoint cycle
& $0.682{\pm}0.009$ & $0.515{\pm}0.008$ \\
& \textbf{+ Trajectory cycle}
& $\mathbf{0.657{\pm}0.008}$ & $\mathbf{0.549{\pm}0.007}$ \\
\midrule
\multirow{5}{*}{0.5}
& Marginal matching only
& $0.706{\pm}0.010$ & $0.476{\pm}0.010$ \\
& + Endpoint cycle
& $0.682{\pm}0.009$ & $0.515{\pm}0.008$ \\
& + Trajectory cycle
& $0.657{\pm}0.008$ & $0.549{\pm}0.007$ \\
& Paired-only
& $0.660{\pm}0.008$ & $0.512{\pm}0.004$ \\
& \textbf{Semi-paired (ours)}
& $\mathbf{0.559{\pm}0.007}$ & $\mathbf{0.638{\pm}0.004}$ \\
\midrule
\multirow{5}{*}{1.0}
& Marginal matching only
& $0.706{\pm}0.010$ & $0.476{\pm}0.010$ \\
& + Endpoint cycle
& $0.682{\pm}0.009$ & $0.515{\pm}0.008$ \\
& + Trajectory cycle
& $0.657{\pm}0.008$ & $0.549{\pm}0.007$ \\
& Paired-only
& $0.508{\pm}0.005$ & $0.664{\pm}0.002$ \\
& \textbf{Semi-paired (ours)}
& $\mathbf{0.481{\pm}0.003}$ & $\mathbf{0.677{\pm}0.002}$ \\
\bottomrule
\end{tabular}
\end{table*}

\newpage
\section{Additional Experiments}
We include an additional comparison to LADB~\citep{wang2025ladb} on the
depth-to-image translation benchmark, which is the closest semi-supervised latent diffusion bridge setting to ours. This comparison is intended to complement the main LDDBM~\cite{berman2025towards} and DiWa~\cite{moser2024waving} comparisons: LDDBM isolates the effect of adding structural constraints to a paired latent bridge, DiWa provides a task-specific SR reference, and LADB tests against a recent method designed to reduce pairing requirements. As shown in Table~\ref{tab:depth2img_ladb_comparison_copy}, \methodname{} remains competitive or better across the evaluated paired fractions, supporting the claim that directly regularizing the bridge geometry is complementary to reducing pairing through latent-alignment baselines.

\begin{table*}[!h]
\centering
\small
\setlength{\tabcolsep}{4.5pt}
\renewcommand{\arraystretch}{1.15}
\caption{\textbf{Comparison with LADB \cite{wang2025ladb} under reduced paired supervision.} \textbf{SDB} is compared to a recent semi-supervised latent diffusion bridge baseline under matched paired-fraction settings.}
\label{tab:depth2img_ladb_comparison_copy}
\resizebox{\textwidth}{!}{%
\begin{tabular}{lcccccccccccccccc}
\toprule
& \multicolumn{4}{c}{$\rho=0$}
& \multicolumn{4}{c}{$\rho=0.25$}
& \multicolumn{4}{c}{$\rho=0.50$}
& \multicolumn{4}{c}{$\rho=1.00$} \\
\cmidrule(lr){2-5}\cmidrule(lr){6-9}\cmidrule(lr){10-13}\cmidrule(lr){14-17}
Method
& FID$\downarrow$ & IS$\uparrow$ & LPIPS$\downarrow$ & MSE$\downarrow$
& FID$\downarrow$ & IS$\uparrow$ & LPIPS$\downarrow$ & MSE$\downarrow$
& FID$\downarrow$ & IS$\uparrow$ & LPIPS$\downarrow$ & MSE$\downarrow$
& FID$\downarrow$ & IS$\uparrow$ & LPIPS$\downarrow$ & MSE$\downarrow$ \\
\midrule
LADB
& -- & -- & -- & --
& 33.44 & 2.29 & 0.6375 & 0.1195
& 33.78 & 2.35 & 0.6335 & 0.1125
& 34.78 & 2.43 & 0.6387 & 0.1052 \\
SDB
& 35.2 & 2.18 & 0.666 & 0.125
& 32.9 & 2.31 & 0.629 & 0.116
& 32.8 & 2.38 & 0.623 & 0.110
& 33.9 & 2.46 & 0.628 & 0.103 \\
\bottomrule
\end{tabular}}
\end{table*}
\end{document}